\documentclass[10pt,journal,compsoc]{IEEEtran}
\usepackage{graphicx}
\usepackage{multirow}
\usepackage{makecell}
\usepackage{mathrsfs}
\usepackage{bbding}
\usepackage{amsmath}
\usepackage{threeparttable}
\usepackage{color}
\usepackage{subfigure}
\usepackage{ragged2e}
\usepackage{amssymb}

\ifCLASSINFOpdf
\else
\fi
\begin{document}
%
\title{A Comprehensive Survey of Scene Graphs: Generation and Application}
%
%
%
%

\author{Xiaojun~Chang,
        Pengzhen~Ren,
        Pengfei~Xu,
        Zhihui~Li,
        Xiaojiang~Chen,
        and~Alex~Hauptmann
\IEEEcompsocitemizethanks{\IEEEcompsocthanksitem X. Chang is with the ReLER Lab, AAII, Faculty of Engineering and Information Technology, University of Technology Sydney, Australia. E-mail: Xiaojun.Chang@uts.edu.au.

\IEEEcompsocthanksitem P. Ren, P. Xu, and X. Chen are with the School of Information Science and Technology, Northwest University.

\IEEEcompsocthanksitem Z. Li is with Qilu University of Technology (Shandong Academy of Sciences), Shandong Artificial Intelligence Institute.
\IEEEcompsocthanksitem A. Hauptmann is with the School of Computer Science, Carnegie Mellon University.}
\thanks{Corresponding authors: Pengfei Xu (pfxu@nwu.edu.cn) and Zhihui Li (zhihuilics@gmail.com).}
\thanks{Manuscript received Mar. 7 2021; revised Aug. 5 2021; revised Sep. 20 2021; accepted Dec. 20 2021.}}

\markboth{IEEE Transactions on Pattern Analysis and Machine Intelligence}%
{Chang \MakeLowercase{\textit{et al.}}: Scene Graph: A Review of Generation and Application}
%



\IEEEtitleabstractindextext{
\begin{abstract}
\justifying
Scene graph is a structured representation of a scene that can clearly express the objects, attributes, and relationships between objects in the scene. As computer vision technology continues to develop, people are no longer satisfied with simply detecting and recognizing objects in images; instead, people look forward to a higher level of understanding and reasoning about visual scenes. For example, given an image, we want to not only detect and recognize objects in the image, but also understand the relationship between objects (visual relationship detection), and generate a text description (image captioning) based on the image content. Alternatively, we might want the machine to tell us what the little girl in the image is doing (Visual Question Answering (VQA)), or even remove the dog from the image and find similar images (image editing and retrieval), etc. These tasks require a higher level of understanding and reasoning for image vision tasks. The scene graph is just such a powerful tool for scene understanding. Therefore, scene graphs have attracted the attention of a large number of researchers, and related research is often cross-modal, complex, and rapidly developing. However, no relatively systematic survey of scene graphs exists at present. To this end, this survey conducts a comprehensive investigation of the current scene graph research. More specifically, we first summarize the general definition of the scene graph, then conducte a comprehensive and systematic discussion on the generation method of the scene graph (SGG) and the SGG with the aid of prior knowledge. We then investigate the main applications of scene graphs and summarize the most commonly used datasets. Finally, we provide some insights into the future development of scene graphs.
\end{abstract}

}

\maketitle

\IEEEdisplaynontitleabstractindextext

%
\IEEEpeerreviewmaketitle

\IEEEraisesectionheading{\section{Introduction}\label{sec:introduction}}

\IEEEPARstart{T}{o} date, deep learning \cite{liang2017deep,ren2020comprehensive,schuster2015generating,ren2020survey,YanCLGGZZ21} has substantially promoted the development of computer vision. People are no longer satisfied with simple visual understanding tasks such as object detection and recognition. Higher-level visual understanding and reasoning tasks are often required to capture the relationship between objects in the scene as a driving force. The emergence of scene graphs is precisely to solve this problem. Scene graphs were first proposed \cite{johnson2015image} as a data structure that describes the object instances in a scene and the relationships between these objects. A complete scene graph can represent the detailed semantics of a dataset of scenes, but not a single image or a video; moreover, it contains powerful representations that encode 2D/3D images \cite{johnson2015image,armeni20193d} and videos \cite{qi2018scene,wangstorytelling} into their abstract semantic elements without restricting either the types and attributes of objects or the relationships between objects. 
Related research into scene graphs greatly promotes the understanding of various tasks such as vision, natural language, and their cross-domains.

As early as 2015, the idea of utilizing the visual features of different objects contained in the image and the relationships between them was proposed as a means of achieving a number of visual tasks, including action recognition \cite{aksoy2010categorizing}, image captioning \cite{aditya2015images} and other relevant computer vision tasks \cite{schuster2015generating}. Visual relationship mining has been demonstrated to significantly improve the performance of related visual tasks, as well as to effectively enhance people's ability to understand and reason about visual scenes. Subsequently, visual relationship was incorporated into scene graph theory by Johnson et al. \cite{johnson2015image}, in which the definition of scene graphs was formally provided. In \cite{johnson2015image}, a scene graph is generated manually from a dataset of real-world scene graphs, enabling the detailed semantics of a scene to be captured. Since then, the research on scene graphs has received extensive attention \cite{tripathi2021sg2caps, rosinol2021kimera, zhang2019graphical}.

\begin{figure} 
  \centering
  \includegraphics[width=0.45\textwidth]{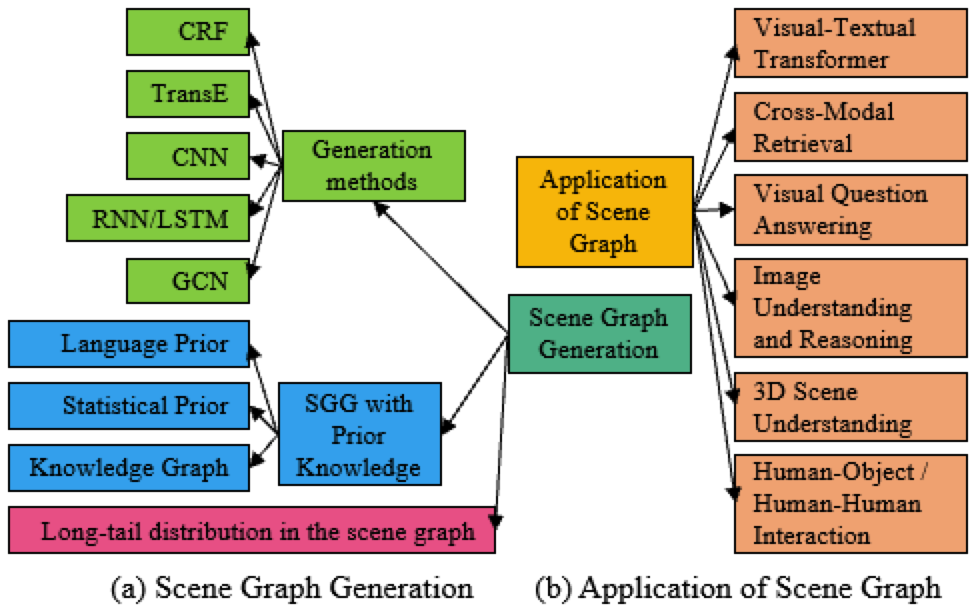}
  \vspace{-1 em}
  \caption{A classification of the methods and applications of SGG.}
  \vspace{-1.5 em}
  \label{fig:SGG_statistics}
\end{figure}

At present, the work related to scene graph generation (SGG) is explosively increasing, but there is a lack of a comprehensive and systematic survey of SGG. In order to fill this gap, we will mainly review the methods and applications of SGG. Fig.~\ref{fig:SGG_statistics} shows the main structure of our survey. In addition, in Section \ref{sec:Datasets and performance evaluation}, we summarize the datasets and evaluation methods commonly used in scene graphs and compare the performance of the models. In Section \ref{sec:FutureDirection}, we discuss the future direction of SGG. Finally, we present our concluding remarks in Section \ref{sec:Conclusion}.

\vspace{-1em}
\subsection{Definition}
\begin{figure*} 
	\centering
	\includegraphics[width=0.9\linewidth]{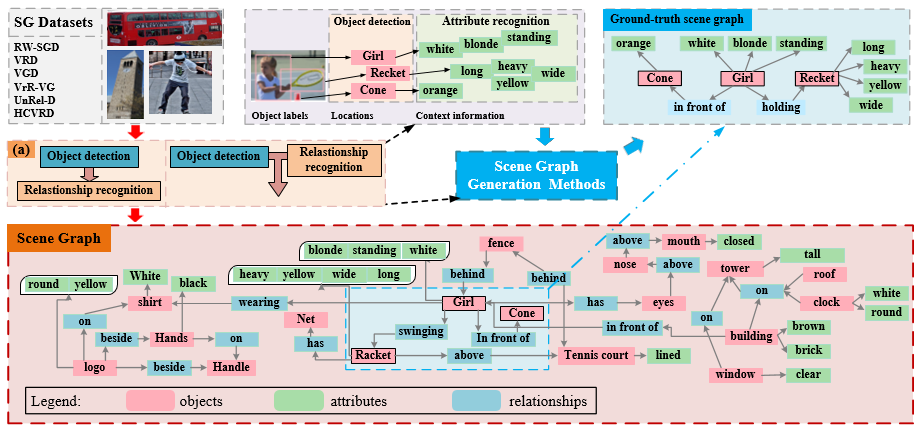}
	\vspace{-1.5 em}
	\caption{An example of scene graph construction. Upper right: The ground-truth scene graph of a given image. Bottom: An example of a complete scene graph.}
	\vspace{-2 em}
	\label{fig:SGG_Example}
\end{figure*}
Fig.~\ref{fig:SGG_Example} summarizes the overall process of building a scene graph. As shown in Fig.~\ref{fig:SGG_Example} (bottom), an object instance in the scene graph can be a person (girl), a place (tennis court), a thing (shirt), or parts of other objects (arms). Attributes are used to describe the state of the current object; these may include its shape (a racket is a long strip), color (girl's clothes are white), and pose (a girl who is standing). Relations are used to describe the connections between pairs of objects, such as actions (e.g., girl swinging racket), and positions (cone placed in front of a girl). This relationship is usually expressed as a $\langle subject-predicate-object \rangle$ triplet, \textcolor{black}{which can be abbreviated as $\langle s-p-o \rangle$}.

Formally, the scene graph $SG$ is a directed graph data structure; this can be defined in the form of a tuple $SG = (O,R,E)$, where $O = \{o_{1}, ...,o_{n}\}$ is the set of objects detected in the images, \textcolor{black}{and $n$ is the number of objects}. Each object can be denoted as $o_{i} = (c_{i},A_{i})$, where $c_{i}$ and $A_{i}$ represent the category and attributes of the object respectively. $R$ stands for a set of relationships between the nodes, where the relationship between the $i$-th and the $j$-th object instance can be expressed as $r_{i\rightarrow j}$,  \textcolor{black}{$i,j\in \left \{ 1,2,...,n \right \}$}. $E\subseteq O\times R\times O$ represents the edges between the object instance nodes and the relationship nodes, \textcolor{black}{so there are at most $n\times n$ edges in the initial graph. Then, $Edge(o_i,r_{i\rightarrow j})\in E$ is }automatically removed when $o_i$ is classified as background or $r_{i\rightarrow j}$ is classified as irrelevant. That is, given an image $I$ as an input, the SGG method outputs a scene graph $SG$, which contains the object instances localized in the image by the bounding boxes and the relationships between each pair of object instances. It can be expressed as follows:
\begin{equation}
    SG_{O,R,E}^I = SGG(I).
\end{equation}
The scene graph can abstract objects and the relationships between them in a more explicit and easier-to-understand way. Therefore, it is regarded as a powerful tool for visual scene understanding.

\begin{figure} 
  \centering
  \includegraphics[width=0.49\textwidth]{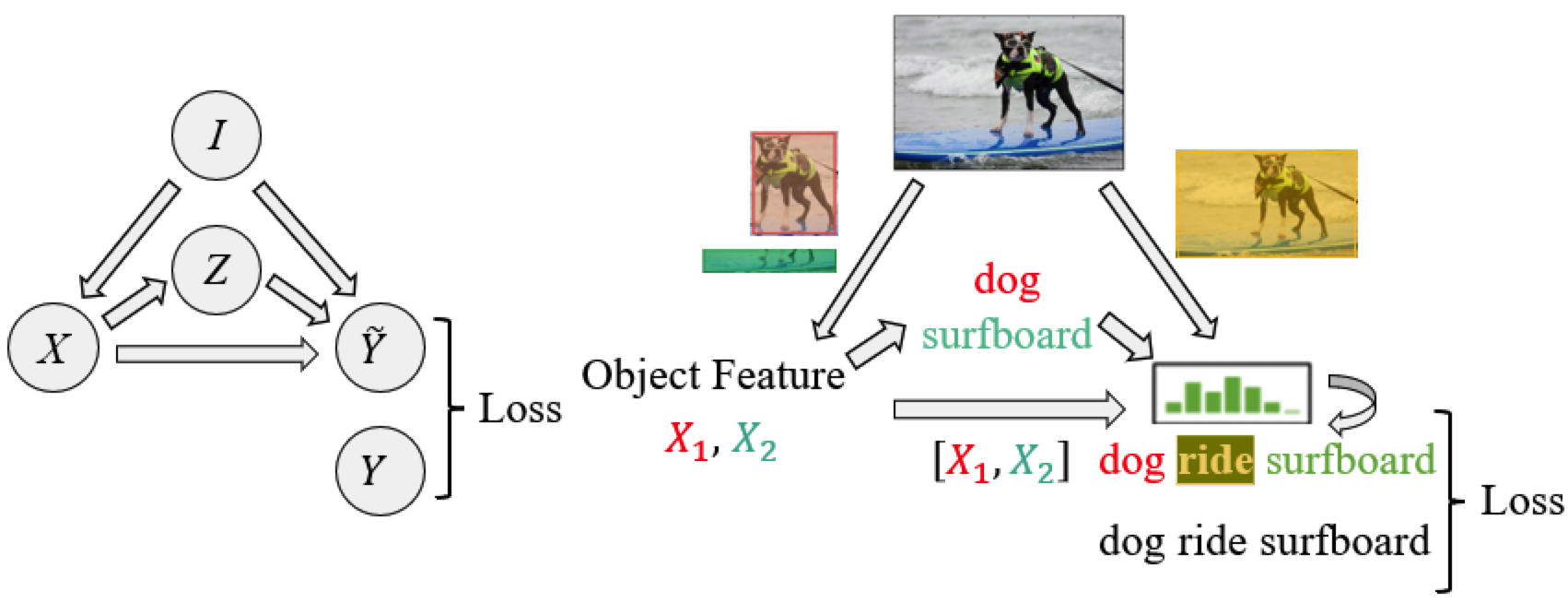}
  \vspace{-2 em}
  \caption{The causal graph of SGG. Left: An abstract representation of a general SGG process. Right: An example of a general SGG process.}
  \vspace{-1.5 em}
  \label{fig:dog_ride_surfboard}
\end{figure} 

\vspace{-1em}
\subsection{Construction Process}
\label{sec:Construction process}
\textcolor{black}{
Referring to the expression in \cite{Tang2020}, a general SGG process is shown in Fig.~\ref{fig:dog_ride_surfboard}. Fig.~\ref{fig:dog_ride_surfboard}(left) shows an abstract representation of this SGG process, and Fig.~\ref{fig:dog_ride_surfboard}(right) shows a concrete example. Specifically, node $I$ represents a given image, node $X$ represents the feature of the object, and node $Z$ represents the category of the object. The node $Y$ represents the category of the prediction predicate and the corresponding triple $\langle s-p-o \rangle$, which uses a fusion function to receive the output from the three branches to generate the final logits. The node $Y$ represents the true triplet label. The explanation of the corresponding link is as follows:\\
\textbf{Link $I\rightarrow X$ (Object Feature Extraction).} A pre-trained Faster R-CNN \cite{Ren2017Faster} is often used to extract a set of bounding boxes $B=\{b_i|i=1,...,m\}$ and the corresponding feature maps $X=\{x_i|i=1,...,m\}$ of the input image $I$. This process can be expressed as:
\begin{equation}
    Input:\{I\}\Rightarrow Output:\{x_i|i=1,...,m\}.
\end{equation}
Through this process, the visual context is encoded for each object.\\
\textbf{Link $X\rightarrow Z$ (Object Classification).} This process can be simply expressed as:
\begin{equation}
    Input:\{x_i\}\Rightarrow Output:\{z_i, z_i\in O\}, i=1,...,m.
\end{equation}
\textbf{Link $Z\rightarrow \widetilde{Y}$ (Object Class Input for SGG).} The paired object label $(z_i,z_j)$ is used, and the predicate $\widetilde{y}_{ij}$ between the object pair is predicted by a combined embedding layer $M$. This process can be expressed as:
\begin{equation}
    Input:\{(z_i,z_j)\} \overset{M}{\Longrightarrow} Output:\{\widetilde{y}_{ij}\}, i\neq j; i,j=1,...,m.
\end{equation}
Some prior knowledge (e.g., language prior \cite{liang2017deep, cui2018context, lu2016visual} and statistical prior \cite{yu2017visual, chen2019knowledge, dai2017detecting}) can be calculated in this link.\\
\textbf{Link $X\rightarrow \widetilde{Y}$ (Object Feature Input for SGG).} The combination $[x_i,x_j]$ of paired object features is used as input to predict the corresponding predicate. This process can be expressed as:
\begin{equation}
    Input:\{[x_i,x_j]\} \Rightarrow Output:\{\widetilde{y}_{ij}\}, i\neq j; i,j=1,...,m.
\end{equation} The rich context information can be fully excavated in this link \cite{newell2017pixels, yu2017visual, yin2018zoom, zhang2018interpretable}.\\
\textbf{Link $I\rightarrow \widetilde{Y}$ (Visual Context Input for SGG).} The visual context feature $v_{ij}=Convs(RoIAlign(I,b_i\bigcup b_j))$ \cite{he2017mask} of the joint region $b_i\bigcup b_j$ is extracted in this link and predicts the corresponding triplet. This process can be expressed as:
\begin{equation}
    Input:\{v_{ij}\} \Rightarrow Output:\{\widetilde{y}_{ij}\}, i\neq j; i,j=1,...,m.
\end{equation}
\textbf{Training Loss.} Most of the models are trained by using the conventional cross-entropy losses of the object label and the predicate label. In addition, to avoid any single link spontaneously dominating
the generation of logits $\widetilde{Y}$, \cite{Tang2020} add auxiliary cross-entropy losses that individually predict
$\widetilde{Y}$ from each branch. Further, \cite{suhail2021energy} improve the training loss from the perspective of the structure of the output space.}

\textcolor{black}{More concise, the generation of scene graphs can be roughly divided into three parts: feature extraction, contextualization, and graph construction and reasoning.
\begin{itemize}
    \item \textit{Feature extraction}. This process is mainly responsible for encoding objects or object pairs in the image. For example, Link $I\rightarrow X$ and Link $I\rightarrow \widetilde{Y}$.
    \item \textit{Contextualization}. It plays the role of associating different entities and is mainly used to enhance the contextual information between entities. For example, Link $I\rightarrow X$, Link $Z\rightarrow \widetilde{Y}$, Link $X\rightarrow \widetilde{Y}$ and Link $I\rightarrow \widetilde{Y}$.
    \item \textit{Graph construction and inference}. Finally, use these contextual information to predict the predicate and complete the construction and inference of the graph. For example, the prediction of node $\widetilde{Y}$ label.
\end{itemize}}

On the other hand, as shown in Fig. \ref{fig:SGG_Example} (a), from the perspective of the SGG process, the generation of scene graphs can be currently divided into two types \cite{li2018factorizable}. The first approach has two stages, namely object detection and pairwise relationship recognition \cite{dai2017detecting,liao2019Natural,lu2016visual,yu2017visual}. The first stage involved in identifying the categories and attributes of the detected objects is typically achieved using Faster-RCNN \cite{Ren2017Faster}. This method is referred to as the \textit{bottom-up} method.
It can be expressed in the following form:
\begin{equation}
    P(SG|I)=P(B|I)*P(O|B,I)*P(R|B,O,I),
    \label{eq:bottom_up}
\end{equation}
where $P(B|I)$, $P(O|B,I)$ and $P(R|B,O,I)$ represent the bounding box ($B$), object ($O$) and relational ($R$) prediction model, respectively.
The other approach involves jointly detecting and recognizing the objects and their relationships \cite{li2017vip,li2017scene,xu2017scene, liu2021fully}. This method is referred to as the \textit{top-down} method. 
The corresponding probability model can be expressed as:
\begin{equation}
    P(SG|I)=P(B|I)*P(O,R|B,I),
\end{equation}
where $P(O,R|B,I)$ represents the joint inference model of objects and their relationships based on the object region proposals.
At the high level, the inference tasks and other visual tasks involved include recognizing objects, predicting the objects' coordinates, and detecting/recognizing pairwise relationship predicates between objects \cite{woo2018linknet}. Therefore, most current works focus on the key challenge of reasoning the visual relationship.

\vspace{-1em}
\subsection{Challenge}
Notably, however, the research on scene graphs still faces several challenges. At present, scene graph research focuses primarily on trying to solve the following two problems:
\begin{itemize}
    \item \textit{Construction of SGG models.} The key question here is how to build a scene graph step by step. Different learning models have a crucial impact on the accuracy and completeness of the scene graph generated by mining visual text information. Accordingly, the study of related learning models is essential for SGG.
    \item \textit{The introduction of prior knowledge.} In addition to fully mining the objects in the current training set, along with their relationships, some additional prior knowledge is also crucial to the scene graph construction. Another important issue is how to make full use of this existing prior knowledge.
    \item \textit{Long-tailed distribution of visual relationships.} The long-tail distribution of predicates in visual relationship recognition is a key challenge. This long-tailed distribution and the data balance required for model training constitute a pair of inherent contradictions, which will directly affect the model performance.
\end{itemize}

\vspace{-2em}
\section{Scene graph generation}
\label{sec:SGG_Alone}
A scene graph is a topological representation of a scene, the primary goal of which is to encode objects and their relationships. Moreover, the key challenge task is to detect/recognize the relationships between the objects.
Currently, the SGG methods can be roughly divided into CRF-based SGG, TransE-based SGG, CNN-based SGG, RNN/LSTM-based SGG, and GNN-based SGG. In this section, we conduct a detailed review of each category of these methods. 

\begin{figure} 
\center{\includegraphics[width=0.4\textwidth] {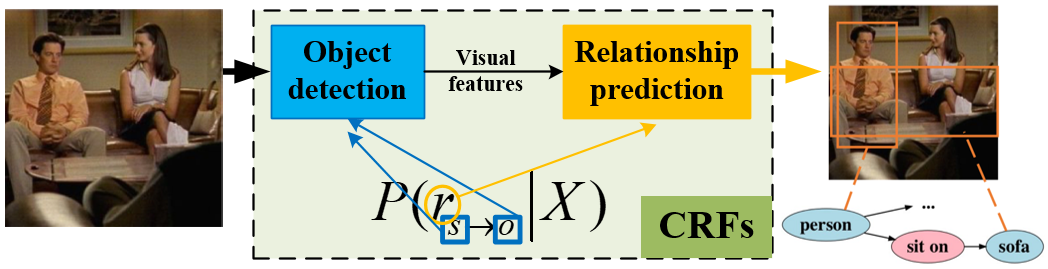}} 
\vspace{-.8 em}
   \caption{The basic structure of CRF-based SGG models has two parts: object detection ($s$ and $o$) and relationship prediction ($r$). }
\vspace{-1.8 em}
 \label{fig:CRF}
\end{figure}

\vspace{-1em}
\subsection{CRF-based SGG}
In the visual relationship triples $\langle s-r-o \rangle$, a strong statistical correlation exists between the relationship predicate and the object pair. Effective use of this information can greatly aid in the recognition of visual relationships. The CRF (Conditional Random Field) \cite{lafferty2001conditional} is a classical tool capable of incorporating statistical relationships into the discrimination task. CRF has been widely used in various graph inference tasks, including image segmentation \cite{krahenbuhl2012efficient, liang2016semantic, zheng2015conditional}, named-entity recognition \cite{lample2016neural, mccallum2003early} and image retrieval \cite{johnson2015image}. In the context of visual relations, the CRF can be expressed as follows:
\begin{equation}
P(r_{s\rightarrow o}|X)=\frac{1}{Z}exp(\Phi(r_{s\rightarrow o}|X;W)), X=x_r,x_s,x_o,
\end{equation}
where $x_r$ refers to the appearance feature and spatial configuration of the object pair, $x_s$ and $x_o$ denote the appearance features of the subject and object respectively. \textcolor{black}{In general, most of these features are one-dimensional tensors after ROI pooling, and have the size of $1\times N$. $N$ is the dimension of tensors, and its specific value can be controlled by network parameters}. $W$ is the parameter of the model, $Z$ is a normalization constant, and $\Phi$ represents the joint potential. Similar CRFs are widely utilized in computer vision tasks \cite{zheng2015conditional, quattoni2004conditional} and have been demonstrated effective in capturing statistical correlations in visual relationships. Fig.\ref{fig:CRF} summarizes the basic structure of CRF-based SGG, which comprises two parts: object detection and relationship prediction. Object detection models are used to obtain the regional visual features of subjects and objects, while relationship prediction models predict the relationships between subjects and objects using their visual features. The other improved CRF-based SGG models achieved better performances by using more suitable object detection models and relation prediction models with stronger inference ability \cite{cong2018scene,dai2017detecting}. For example, Deep Relational Network (DR-Net) and Semantic Compatibility Network (SCN) were proposed in \cite{cong2018scene} and \cite{dai2017detecting}.

Inspired by the success of deep neural networks \cite{simonyan2014very, he2016deep} and CRF \cite{lafferty2001conditional} models, to explore statistical relationships in the context of visual relationships, DR-Net \cite{dai2017detecting} opts to incorporate statistical relationship modeling into the deep neural network framework. DR-Net unrolls the inference of relational modeling into a feedforward network. In addition, DR-Net is different from previous CRFs. More specifically, the statistical inference procedure in DR-Net is embedded in a deep relational network through iteration unrolling. The performance of the improved DR-Net is not only superior to classification-based methods but also better than deep potential-based CRFs. 
Furthermore, SG-CRF (SGG via Conditional Random Fields) \cite{cong2018scene} can be defined as maximizing the following probability function by finding the best prediction of $o_i,o_i^{bbox},r_{i\rightarrow j}$:
\begin{equation}
    P(SG|I)=\prod_{o_i\in O}P(o_i,o_i^{bbox}|I)\prod_{r_{i\rightarrow j\in R}}P(r_{i\rightarrow j}|I),
    \label{eq:EBM_0}
\end{equation}
where $o_i^{bbox} \in\mathbb{R}^4$ represents the bounding box coordinate of the $i$-th object instance. It has been observed that some previous methods \cite{krishna2017visual, lu2016visual, Sadeghi2011Recognition, xu2017scene} tend to ignore the semantic compatibility (that is, the likelihood distribution of all 1-hop neighbor nodes of a given node) between instances and relationships, which results in a significant decrease in the model performance when faced with real-world data. For example, this may cause the model to incorrectly recognize $\langle dog-sitting~inside-car \rangle$ as $\langle dog-driving-car \rangle$. Moreover, these models ignore the order of the two, leading to confusion between subject and object, which may produce absurd predictions such as $\langle car-sitting~inside-dog \rangle$. To solve these problems, an end-to-end scene graph constructed via conditional random fields was proposed by SG-CRF to improve the quality of SGG. More specifically, to learn the semantic compatibility of nodes in the scene graph, SG-CRF proposes a new semantic compatibility network based on conditional random fields. To distinguish between the subject and object in the relationship, SG-CRF proposes an effective relation sequence layer that can capture the subject and object sequence in the visual relationship. 

In general, the CRF-based SGG can effectively model the statistical correlation in the visual relationship. This statistically relevant information modeling remains a classic tool in visual relationship recognition tasks.

\begin{figure} 
    \center{\includegraphics[width=0.49\textwidth] {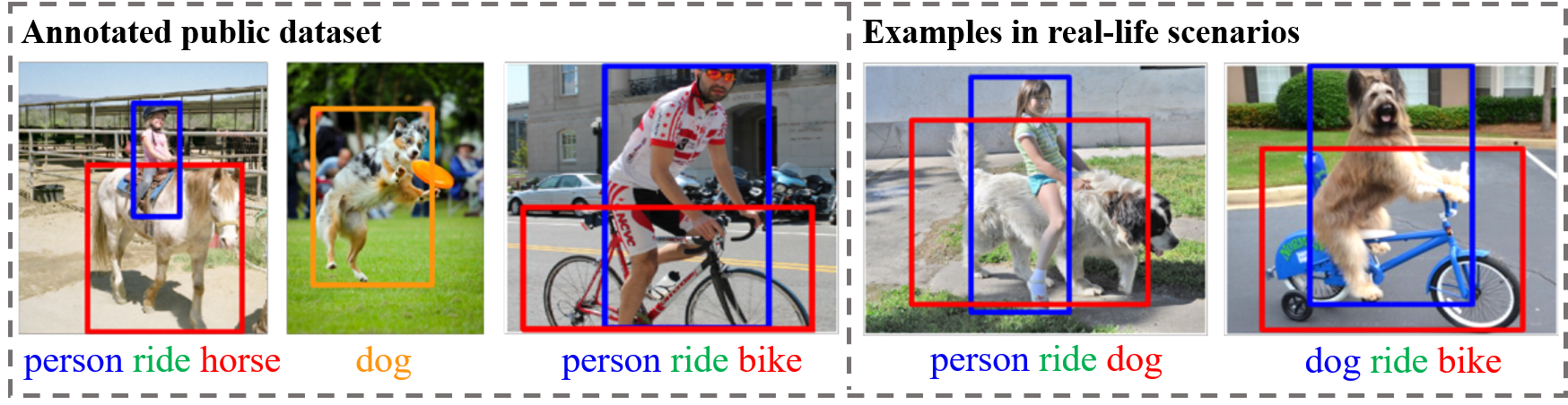}} 
    \vspace{-1.8 em}
    \caption{Examples of the sparsity and variability of visual relationships. 
    }
    \vspace{-1.2 em}
    \label{fig:unsee_relation}
\end{figure}

\vspace{-1em}
\subsection{TransE-based SGG}
\label{sec:TransE-based SGG}

The knowledge graph is similar to the scene graph; it also has a large number of fact triples, and these multi-relational data are denoted in the form $(head,label,tail)$ (abbreviated as $(h,l,t)$). Among them, $h,t\in O$ are the head entity and the tail entity respectively, while $l$ is the relationship label between the two entities. \textcolor{black}{These entities are the objects in scene graph, so we use $O$ to denote the set of entities for avoiding confusion with $E$ (the edges between the objects)}. Knowledge graphs represent learning to embed triples into low-dimensional vector spaces, with TransE (Translation Embedding)-based \cite{Bordes2013Translating} models having been demonstrated particularly effective. Furthermore, TransE regards the relationship as a translation between the head entity and the tail entity. The model is required to learn vector embeddings of entities and relationships. That is, for the tuple $(h,l,t)$, $h+l\approx t$ ($t$ should be the nearest neighbor of $h+l$; otherwise, they should be far away from each other). This embedding learning can be achieved by minimizing the following margin-based loss function:
\begin{equation}
\mathcal{L}=\sum_{(h,l,t)\in S}\sum_{(h',l,t')\in S'_{(h,l,t)}}[\gamma+d(h+l,t)-d(h'+l,t')]_+,
\end{equation}
where $S$ represents the training set, $d$ is used to measure the distance between the two embeddings, $\gamma$ is a marginal hyperparameter, $(h,l,t)$ is a positive tuple, $(h',l,t')$ is a negative tuple, and
\begin{equation}
S'_{h,l,t}=\{(h',l,t)|h'\in \mathbb{E}\}\cup \{(h,l,t')|t'\in O\}.
\end{equation}
The relationship tuples in the scene graph also have similar definitions and attributes, meaning that the learning of this visual relationship embedding is also very helpful for the scene graph. 

\begin{figure} 
\center{\includegraphics[width=0.4\textwidth] {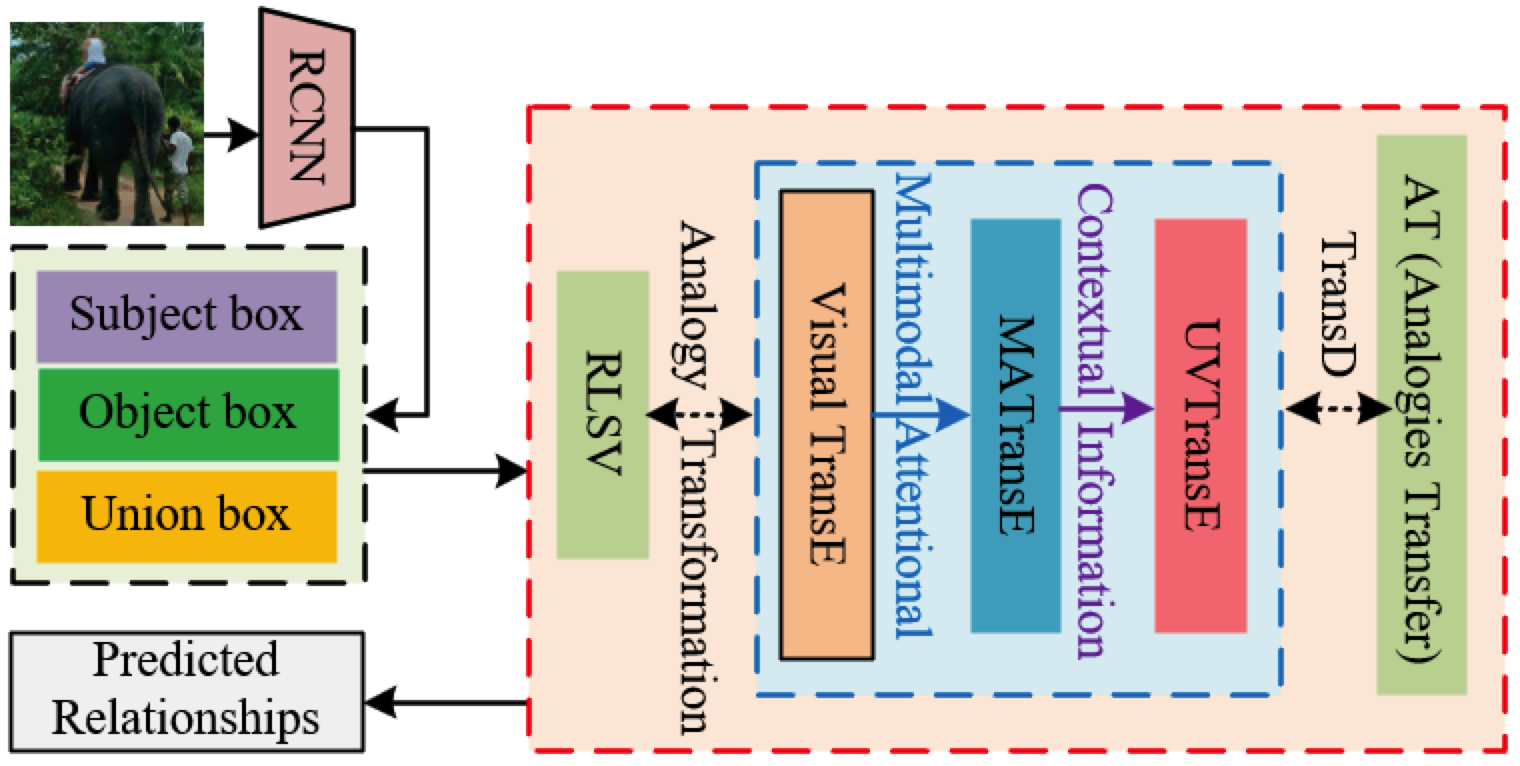}} 
\vspace{-.8 em}
   \caption{Relevant TransE-based SGG models.}
\vspace{-1.5 em}
 \label{TransE}
\end{figure}

Inspired by the advances made by TransE in the relational representation learning of knowledge bases \cite{Bordes2013Translating,lin2015learning}, VTransE \cite{zhang2017visual} (based on TransE) explored how visual relations could be modeled by mapping the features of objects and predicates in low-dimensional space, and is the first TransE-based SGG method that works by extending TransE networks \cite{Bordes2013Translating}. Subsequently, as shown in Fig. \ref{TransE}, attention mechanisms and visual context information are introduced for designing MATransE (Multimodal Attentional Translation Embeddings) \cite{gkanatsios2019deeply} and UVTransE (Union Visual Translation Embedding) \cite{hung2019union} respectively. Furthermore, TransD and analogy transformation are used to replace TransE for visual relationship prediction in RLSV (Representation Learning via Jointly Structural and Visual Embedding)  \cite{ji2015knowledge} and AT (Analogies Transfer) \cite{peyre2019detecting} respectively.

More specifically, VTransE maps entities and predicates into a low-dimensional embedding vector space, in which the predicate is interpreted as the translation vector between the embedded features of the subject and the object's bounding box regions. \textcolor{black}{Similar to the tuple of knowledge graph, the relationship in scene graph is modeled as a simple vector transformation, i.e., $s + p \approx o$.}
It can be consided as a basic vector transformation for TransE-based SGG methods. While this is a good start, VTransE considers only the features of the subject and the object, and not those of the predicate and context information \cite{ren2020scene, jiang2021learning}, despite these having been demonstrated to be useful for the recognition of relations \cite{zhuang2017towards, yin2018zoom}. To this end, MATransE \cite{gkanatsios2019deeply}, an approach based on VTransE, combines the complementary nature of language and vision \cite{lu2016visual}, along with an attention mechanism \cite{jetley2018learn} and deep supervision \cite{xie2015holistically}, to propose a multimodal attention translation embedding method. \textcolor{black}{MATransE tries to learn the projection matrices $W_s, W_p$ and $W_o$ for a projection of $\langle s, p, o \rangle$ into a score space, and the binary masks’ convolutional features $m$ are used in the attention module. Then, $s + p \approx o$ becomes:} 
\begin{equation}
    W_s(s, o, m)s+W_p (s, o, m)p \approx W_o(s, o, m)o.
\end{equation}
MATransE designed two separate branches to deal directly with those of the predicate and the features of the subject-object, achieving good results.

\begin{figure} 
\center{\includegraphics[width=0.43\textwidth] {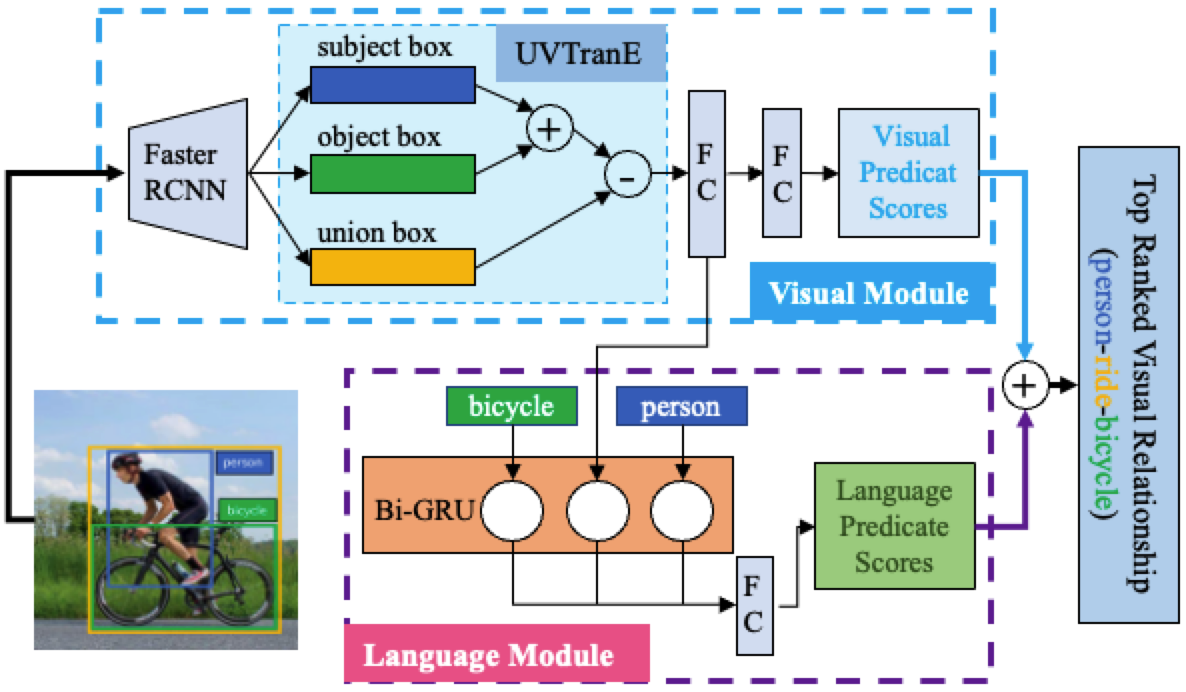}} 
\vspace{-.8 em}
   \caption{The overall structure of UVTransE's \cite{hung2019union} visual detection model.}
\vspace{-1.2 em}
 \label{fig:UVTransE}
\end{figure}

In addition to drastically changing the visual appearance of the predicate, both the sparsity of the predicate representation in the training set \cite{ramanathan2015learning, Sadeghi2011Recognition} and the very large predicate feature space also make the task of visual relationship detection increasingly difficult. 
Let us take the Stanford VRD dataset \cite{lu2016visual} as an example. This dataset contains 100 classes of objects, 70 classes of predicates, and a total of $30k$ training relationship annotations. The number of possible $\langle s-p-o \rangle$ triplets is $100^2*70=700k$, which means that a large number of possible real relationships do not even have a training example. These invisible relationships should not be ignored, even though they are not included in the training set. Fig. \ref{fig:unsee_relation} presents an example of this case.
However, VTransE and MATransE are not well-suited to dealing with this issue. Therefore, the detection of unseen/new relationships in scenes is essential to the building of a complete scene graph. Inspired by VTransE \cite{zhang2017visual}, the goal of UVTransE \cite{hung2019union} is to improve the generalization for rare or unseen relations.
Based on VTransE, UVTransE introduces a joint bounding box \textcolor{black}{or union feature $u$} of subject and object to facilitate better capturing of contextual information and learns the embeddings \textcolor{black}{ of the predicate} guided by the constraint $p \approx u-s-o$.
UVTransE introduces the union of subject and object and uses a context-augmented translation embedding model to capture both common and rare relations in scenes. This type of exploration is highly beneficial for constructing a relatively complete scene graph. Finally, UVTransE combines the scores of vision, language, and object detection modules to sort the predictions of the triple relationship. 
The architectural details of UVTRansE are illustrated in Fig. \ref{fig:UVTransE}. UVTransE treats predicate embedding as $p \approx u(s, o)-(s+o)$. While, VTransE \cite{zhang2017visual} models visual relationships by mapping the features of objects and predicates in a low-dimensional space, where the relationship triples can be interpreted as vector translation: $s+p\approx o$.

In addition, to solve the same problem associated with new relationship discovery in scenes to generate incomplete scene graphs, RLSV \cite{wan2018representation} attempts to use existing scene graphs and images to predict the new relationship between two entities, enabling it to achieve scene graph completion. RLSV begins with the relevant knowledge of the knowledge graph, incorporating the characteristics of the scene graph, and proposes an end-to-end representation learning model of joint structure and visual embedding. Unlike TransE-based SGG methods, RLSV uses TransD \cite{ji2015knowledge} (an improved version of TransR/CTransR \cite{lin2015learning}) to project the entities (subjects and objects) from the entity space to the relation space by means of two mapping matrices \textcolor{black}{($M_s$ and $M_o$), and the triplets $\langle s-p-o \rangle$ have new representations $\langle s\perp-p\perp-o \perp \rangle$. Then, following $s + p \approx o$ of TransE, the relationship in scene graph can be modeled as: $\langle s\perp+ p\perp \approx o \perp \rangle$.} 

Unlike UVTransE and RLSV, which aim to find existing visual relationships but lack corresponding annotations in the image, AT (Analogies Transfer) \cite{peyre2019detecting} tries to detect those visual relationships that are not visible in the training set. As shown in Fig. \ref{fig:unsee_relation}, the individual entities of $\langle person-ride-dog \rangle$ and $\langle dog-ride-bike \rangle$ are available in the training set; however, either their combination is not seen in the training set, or the visual relationship is extremely rare. As is evident, AT studies a more general phenomenon, specifically those unseen relationships that are visible in the training set for a single entity but not for the combination of $\langle s-p-o \rangle$. The whole network model utilizes analogy transformation to compute the similarity between the unseen triplet and its similar triplets in order to estimate this unseen relationship and has achieved good results in unseen relationship detection. \textcolor{black}{Therefore, the transformation learned in AT is to transform the visual phrase embedding $\omega $ of a source triplet $\langle s, p, o\rangle$ to $\omega '$ of a target triplet $\langle s' , p' , o'\rangle$, rather than visual feature space transformations in TransE-based SGG methods.} Compared with the commonly used TransD/TransE-based SGG methods, this SGG method using Analogies Transfer has good research prospects.

Based on the insights obtained from knowledge graph-related research, the TransE-based SGG method has developed rapidly and attracted strong interest from researchers. Related research results have also proven that this method is effective. In particular, the TransE-based SGG method is very helpful for the mining of unseen visual relationships, which will directly affect the integrity of the scene graph. Related research is thus still very valuable.

\begin{figure} 
\center{\includegraphics[width=0.45\textwidth] {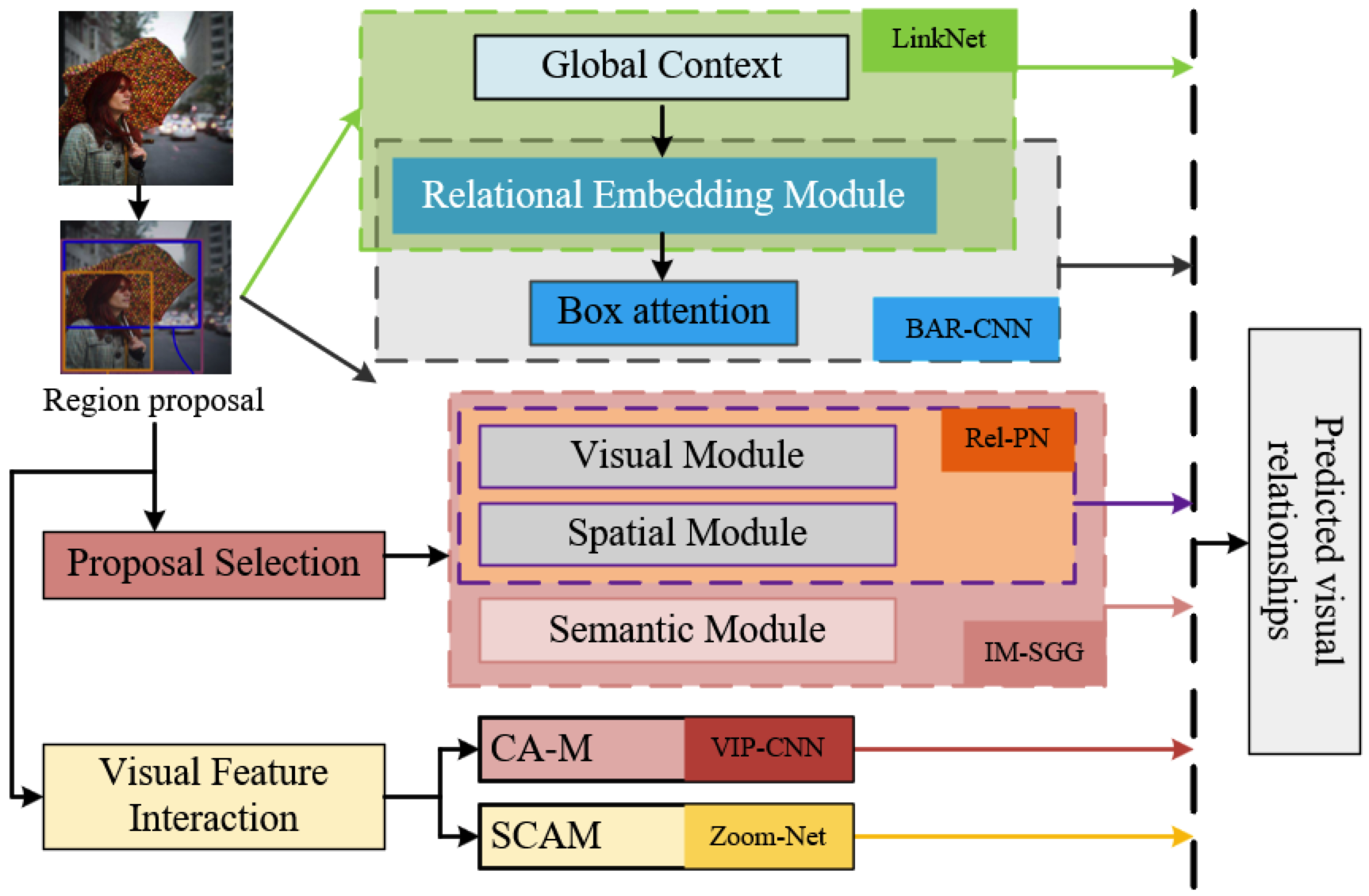}} 
\vspace{-1 em}
   \caption{The mainstream CNN-based SGG models.}
\vspace{-0.5 em}
 \label{fig:CNN}
\end{figure}

\vspace{-1em}
\subsection{CNN-based SGG}
CNN-based SGG methods attempt to extract the local and global visual features of the image using convolutional neural networks (CNN), followed by predicting the relationships between the subjects and objects via classification. \textcolor{black}{From most CNN-based SGG methods, we can conclude that this type of SGG methods mainly includes three parts: region proposal, feature learning and relational classification. Among these parts, feature learning is one of the key parts, and we can use the following formulation to express the feature learning at level $l$ of the subject, predicate and object respectively:
\begin{equation}
   v_{\ast}^{l}=f(W_{\ast}^{l}\otimes v_{\ast}^{l-1}+c_{\ast}^{l}),
   \label{eq:cnn}
\end{equation}
where $\ast$ can be $\left \{s, p, o\right \}$ and $\otimes$ is the matrix-vector product, $W_{\ast}^{l}$ and $c_{\ast}^{l}$ are the parameters of FC or Conv layers. The subsequent CNN-based SGG methods are devoted to designing new modules to learn optimal features $v'$.} 
Fig. \ref{fig:CNN} presents the mainstream CNN-based SGG methods. The final features used for relationship identification are obtained by jointly considering the local visual features of multiple objects in LinkNet \cite{woo2018linknet} or introducing a box attention mechanism in BAR-CNN (Box Attention Relational CNN) \cite{kolesnikov2019detecting}. In an attempt to improve the efficiency of SGG models, Rel-PN \cite{zhang2017relationship} and IM-SGG (Interpretable Model for SGG) \cite{zhang2018interpretable} aim to select the most effective ROIs for visual relational prediction. ViP-CNN (Visual Phrase-guided Convolutional Neural Network) \cite{li2017vip} and Zoom-Net \cite{yin2018zoom} pay more attention to the interactions between local features. As CNN performs well at extracting the visual features of the image, the related SGG method based on CNN has been extensively studied.
In this part, we will elaborate on these CNN-based SGG methods.

The scene graph is generated by analyzing the relationships between multiple objects in the image dataset. It is accordingly necessary to consider the connection between related objects as much as possible, rather than focusing on a single object in isolation. LinkNet \cite{woo2018linknet} was proposed to improve SGG by explicitly modeling inter-dependency among all related objects. More specifically, LinkNet designs a simple and effective relational embedding module that jointly learns the connections between all related objects. In addition, LinkNet also introduces a global context encoding module and a geometrical layout encoding module, which extract global context information and spatial information between object proposals from the entire image and thereby further improve the performance of the algorithm. The specific LinkNet is divided into three main steps: bounding box proposal, object classification, and relationship classification. However, LinkNet considers the relation proposal of all objects, which makes it computationally expensive.
\begin{figure} 
\center{\includegraphics[width=0.49\textwidth] {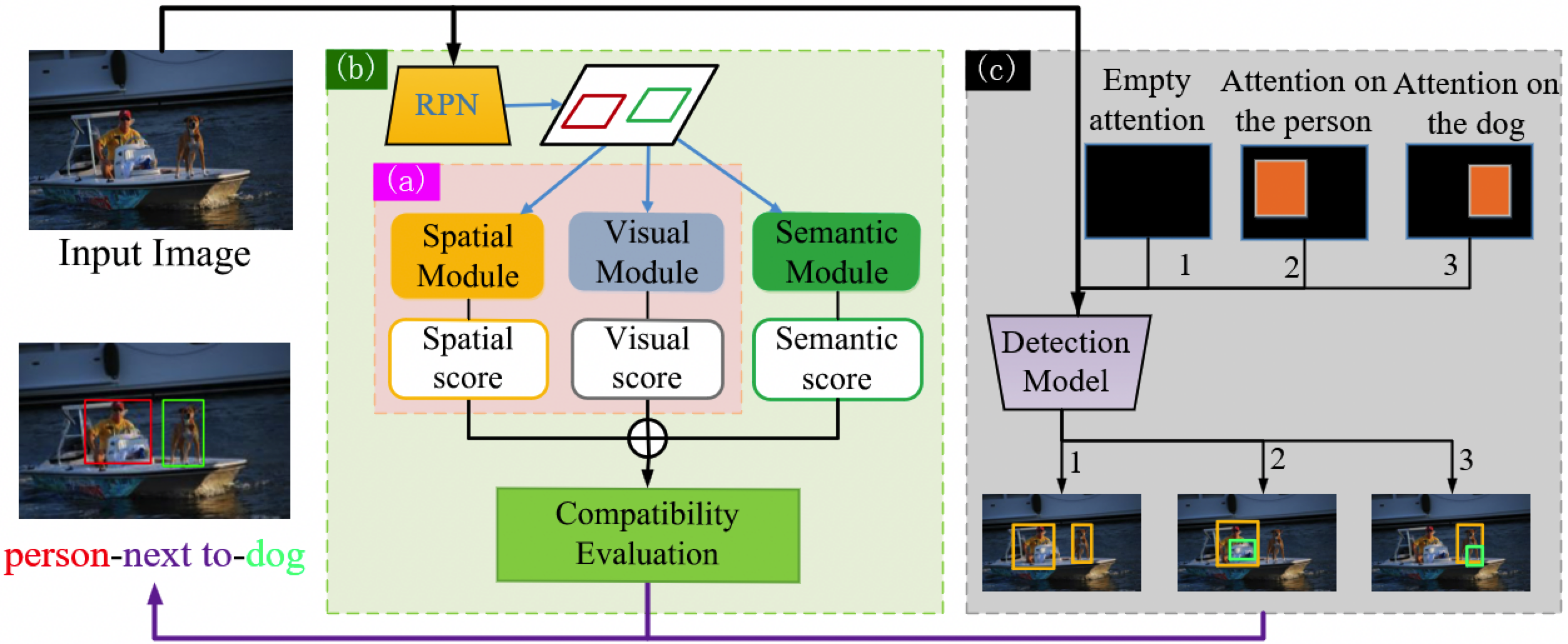}} 
\vspace{-2 em}
   \caption{Comparison of the brief schematic diagrams of three CNN-based SGG methods.}
\vspace{-1.7 em}
   \label{fig:CNN_BASED1}
\end{figure}

On the other hand, as deep learning technology has developed, the corresponding object detection research has become increasingly mature \cite{Ren2017Faster, Redmon2016you, liu2016SSD, kang2017object, kang2018t, kang2016object}. By contrast, the recognition of associations between different entities for higher-level visual task understanding has become a new challenge; this is also the key to scene graph construction. As analyzed in Section \ref{sec:TransE-based SGG}, to detect all relationships, it is inefficient and unnecessary to first detect all single objects and then classify all pairs of relationships, as the visual relationship that exists in the quadratic relationship is very sparse. Using visual phrases \cite{Sadeghi2011Recognition} to express this visual relationship may therefore be a good solution. 
Rel-PN \cite{zhang2017relationship} has conducted corresponding research in this direction. Similar to the region proposals of objects provided by Region Proposal Networks (RPN), Rel-PN \cite{zhang2017relationship} utilizes a proposal selection module to select meaningful subject-object pairs for subsequent relationship prediction. This operation will greatly reduce the computational complexity of SGG. The model structure of Rel-PN is illustrated in Fig. \ref{fig:CNN_BASED1}(a). Rel-PN's compatibility evaluation module uses two types of modules: a visual compatibility module and a spatial compatibility module. The visual compatibility module is mainly used to analyze the coherence of the appearance of the two boxes, while the spatial compatibility module is primarily used to explore the locations and shapes of the two boxes. Furthermore, IM-SGG \cite{zhang2018interpretable}, based on Rel-PN, considers three types of features, namely visual, spatial, and semantic, which are extracted by three corresponding models. Subsequently, similar to Rel-PN, these three types of features are fused for the final relationship identification. Different from Rel-PN, IM-SGG utilizes an additional semantic module to capture the strong prior knowledge of the predicate., and achieve better performance (see Fig. \ref{fig:CNN_BASED1}(b)). This method effectively improves the interpretability of SGG.
More directly, using a similar method to Rel-PN, ViP-CNN \cite{li2017vip} also clearly treats the visual relationship as a visual phrase containing three components. ViP-CNN \cite{li2017vip} attempts to jointly learn the specific visual features for the interaction to facilitate the consideration of the visual dependency. In ViP-CNN, the PMPS (Phrase-guided Message Passing Structure) is proposed to model the interdependency information among local visual features using a gather-broadcast message passing flow mechanism. ViP-CNN has achieved significant improvements in speed and accuracy.

\begin{figure} 
\center{\includegraphics[width=0.35\textwidth] {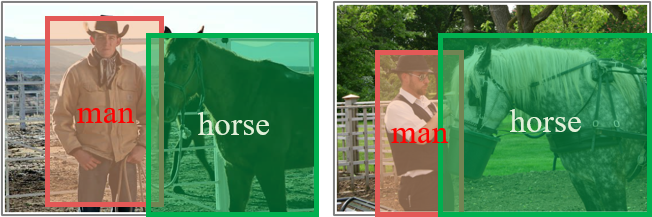}}
\vspace{-.8 em}
   \caption{Schematic diagram of object detection. 
   }
\vspace{-1.8 em}
   \label{fig:object_detection}
\end{figure}

\begin{figure} 
\center{\includegraphics[width=0.49\textwidth]
{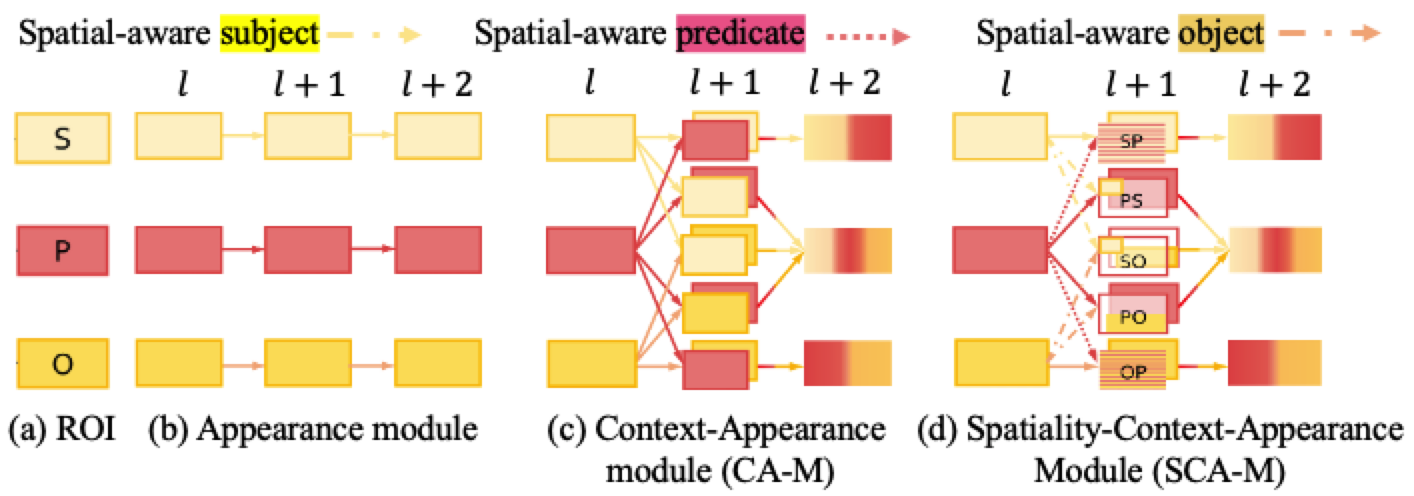}}
\vspace{-2 em}
   \caption{(a) The ROI-pooled feature of the subject (S), predicate (P), and object (O) of a given input image. (b) Appearance Module (A-M) without information interaction. (c) Context-Appearance Module (CA-M) in ViP-CNN \cite{li2017vip}. (d) Spatiality-Context-Appearance Module (SCA-M) in Zoom-Net \cite{yin2018zoom}.}
   \vspace{-1 em}
   \label{fig:ViP-CNN&Zoom-Net}
\end{figure}

In addition, to further improve the SGG accuracy, some methods have also studied the interaction between different features with the goal of more accurately predicting the visual relationship between different entities. This is because the independent detection and recognition of a single object provide little assistance in fundamentally recognizing visual relationships. Fig. \ref{fig:object_detection} presents an example of a case in which even the most perfect object detector finds it difficult to distinguish people standing beside horses from people feeding horses. 
Therefore, the information interaction between different objects is extremely important to the understanding of visual relationships. Many related works have been published on this subject.
For example, the interactions between detected object pairs are used for visual relationship recognition in Zoom-Net \cite{yin2018zoom}. Zoom-Net achieves compelling performance by successfully recognizing complex visual relationships through the use of deep message propagation and the interaction between local object features and global predicate features, without the use of any linguistic priors. VIP-CNN \cite{li2017vip} also uses similar feature interactions. The key difference is that the CA-M (Context-Appearance Module) proposed by VIP-CNN attempts to directly fuse pairwise features to capture contextual information, while the SCA-M (Spatiality-Context-Appearance Module) proposed by Zoom-Net \cite{yin2018zoom} performs spatially-aware channel-level local and global context information fusion. Therefore, SCA-M has more advantages while capturing the spatial and contextual relationships between the subject, predicate, and object features. Fig. \ref{fig:ViP-CNN&Zoom-Net} presents the structure comparison diagram of the Appearance Module (A-M) without information interaction, along with the Context-Appearance Module (CA-M) and Spatiality-Context-Appearance Module (SCA-M).

An attention mechanism is also a good tool for improving visual relationship detection. BAR-CNN \cite{kolesnikov2019detecting} observed that the receptive field of neurons in the most advanced feature extractors \cite{he2016deep,szegedy2016rethinking} may still be limited, meaning that the model may cover the entire attention map. To this end, BAR-CNN proposes a box attention mechanism; this enables visual relationship detection tasks to use existing object detection models to complete the corresponding relationship recognition tasks without introducing additional complex components. This is a very interesting concept, and BAR-CNN has also obtained competitive recognition performance. A schematic illustration of BAR-CNN is presented in Fig. \ref{fig:CNN_BASED1}(c).

Related CNN-based SGG methods have been extensively studied. However, there are still many remaining challenges requiring further research, including those of how to reduce the computational complexity as much as possible while ensuring deep interaction between the triplet's different features, how to deal with the real but very sparse visual relationship in reality, etc. Identifying solutions to these problems will further deepen the research related to the CNN-based SGG method.

\vspace{-1em}
\subsection{RNN/LSTM-based SGG}
\label{sec:RNN/LSTM-based SGG}
\begin{figure} 
\center{\includegraphics[width=0.49\textwidth] {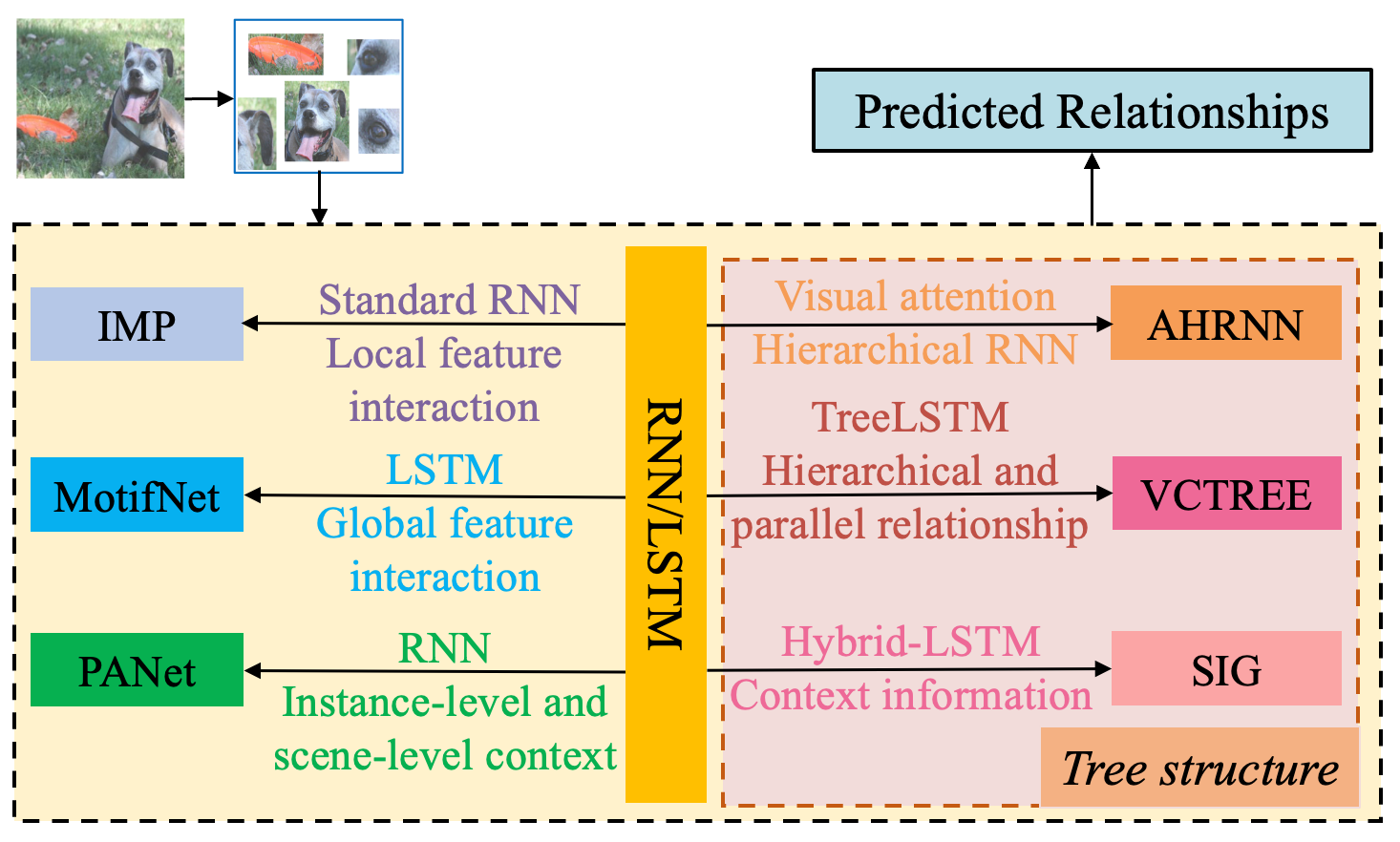}} 
\vspace{-1.8 em}
   \caption{RNN/LSTM-based SGG models.}
\vspace{-1.5 em}
 \label{fig:rnn-lstm}
\end{figure}

A scene graph is a structured representation of an image. The information interaction between different objects and the contextual information of these objects is crucial to the recognition of the visual relationship between them. Models based on RNN and LSTM have natural advantages in capturing the contextual information in the scene graph and reasoning on the structured information in the graph structure. RNN/LSTM-based methods are thus also a popular research direction. As shown in Fig. \ref{fig:rnn-lstm}, based on the standard RNN/LSTM networks, several improved SGG models have been proposed. For example, the feature interaction of local and global context information is considered in IMP (Iterative Message Passing) \cite{xu2017scene} and MotifNet (Stacked Motif Network) \cite{zellers2018neural} respectively. Similarly, instance-level and scene-level context information are used for SGG in PANet (predicate association network) \cite{chen2019panet}, \textcolor{black}{ and attention-based RNN is also introduced in SIG (Sketching Image Gist) \cite{wang2020sketching} for SGG. The corresponding probabilistic interpretation of these RNN/LSTM-based SGG methods can be simplified in the conditional form of Eq.(\ref{eq:bottom_up}), while these methods mainly utilize standard/improved RNN/LSTM networks to inference the relationship by optimizing $P(R|B,O,I)$.} Later, RNN/LSTM-based models have attempted to learn different types of contextual information by designing structural RNN/LSTM modules; examples include AHRNN (attention-based hierarchical RNN) \cite{gao2019hierarchical}, VCTree (Visual Context Tree model) \cite{tang2019learning}. \textcolor{black}{This type of SGG methods considers the scene graph as a hierarchical graphical structure, so they need to construct a hierarchical entity tree based on the region proposals. Then the hierarchy contextal information can be encoded by}
\begin{equation}
   D = BiTreeLSTM(\left \{ z_i \right \}{_{i=0}^{n}}),
    \label{eq:bilstm}
\end{equation}
\noindent \textcolor{black}{where $z_i$ is the feature of the input nodes in the constracted hierarchical entity tree. Finally, a MLP (Multi-layer Perceptron) classifier is used to predict the predicate $p$}.

As discussed above, to make full use of the contextual information in the image to improve the accuracy of SGG, IMP \cite{xu2017scene} was proposed. IMP attempts to use standard RNN to solve the scene graph inference problem and iteratively improve the model's prediction performance through message passing. The main highlight of this approach is its novel primal-dual graph, which enables the bi-directional flow of node information and edge information, and updates the two GRUs \cite{Cho2014} of node and edge in an iterative manner. This form of information interaction helps the model to more accurately identify the visual relationships between objects. 
Unlike cases of interaction between local information, such as IMP, MotifNet \cite{zellers2018neural} begins from the assumption that the strong independence assumption in the local predictor \cite{xu2017scene, li2017vip, li2017scene} limits the quality of global prediction. To this end, MotifNet encodes global context information through recurrent sequential architecture LSTMs (Long Short-term Memory Networks) \cite{hochreiter1997long}.
However, MotifNet \cite{zellers2018neural} only considers the context information between objects while failing to take scene information into account. There have also been some works \cite{xu2017scene, li2018factorizable, yang2018graph} that investigate the classification of relationships by exchanging the context between nodes and edges. 
However, the above-mentioned SGG methods focus primarily on the structural-semantic features in a scene while ignoring the correlations among different predicates. For this reason, \cite{chen2019panet} proposed a two-stage predicate association network (PANet). The main goal of the first stage is to extract instance-level and scene-level context information, while the second stage is mainly used to capture the association between predicate alignment features. In particular, an RNN module is used to fully capture the association between alignment features. This kind of predicate association analysis has also achieved good results.

However, the methods discussed above often rely on object detection and predicate classification between objects. There are two inherent limitations of this approach: first, the object bounding box or relationship pairs generated via the object detection method are not always necessary for the generation of the scene graph; second, SGG depends on the probabilistic ranking of the output relationships, which will lead to semantically redundant relationships \cite{masui2017recurrent}.
For this reason, AHRNN \cite{gao2019hierarchical} proposed a hierarchical recurrent neural network based on a visual attention mechanism. This approach first uses the visual attention mechanism \cite{mnih2014recurrent, nguyen2013image} to resolve the first limitation. Secondly, AHRNN regards the recognition of relational triples as a sequence learning problem using recurrent neural networks (RNN). In particular, it employs hierarchical RNN to model relational triples to more effectively process long-term context information and sequence information \cite{li2015hierarchical, lin2015hierarchical}, thereby avoiding the need to rank the probability of output relationships.

\begin{figure} 
\center{\includegraphics[width=0.49\textwidth] {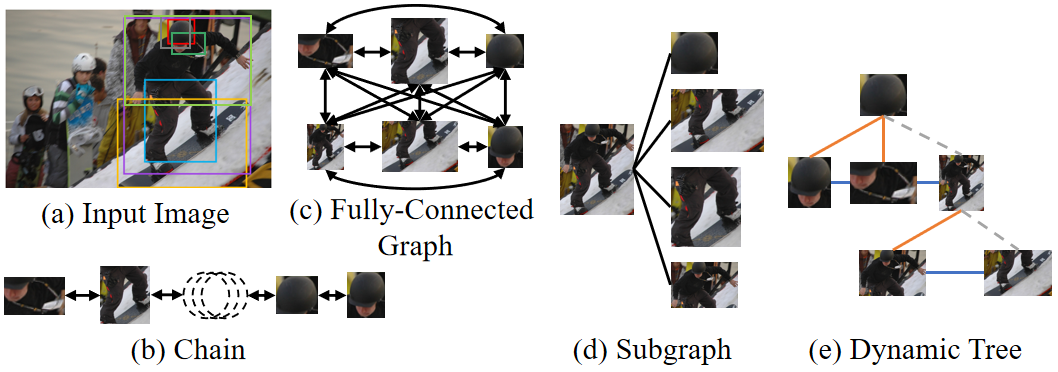}} 
\vspace{-1.5 em}
   \caption{A comparison of the chains \cite{zellers2018neural}, fully connected graph \cite{xu2017scene, chen2018scene, chen2018iterative, dai2017detecting, li2018factorizable, xia2016weakly, yin2018zoom}, subgraph \cite{li2018factorizable} and dynamic tree \cite{tang2019learning} structure of the scene graph. The dynamic tree structure on the left shows a left-child right-sibling binary tree, where the left branches (red) represent the hierarchical relationships, while the right branches (blue) represent the parallel relationship. 
   }
   \vspace{-1.8 em}
   \label{fig:Chain_Graph_VCTree}
\end{figure}

On the other hand, VCTree \cite{tang2019learning} observed that the previous scene graphs either adopted chains \cite{zellers2018neural} or a fully-connected graph \cite{xu2017scene, chen2018scene, chen2018iterative, dai2017detecting, li2018factorizable, xia2016weakly, yin2018zoom}. However, VCTree proposes that these two prior structures may not be optimal, as the chain structure is too simple and may only capture simple spatial information or co-occurrence bias; moreover, the fully connected graph lacks the distinguishing structure of hierarchical and parallel relationships. To solve this problem, VCTree proposed composite dynamic tree structures, which can use TreeLSTM \cite{tai2015improved} for efficient context coding and thus effectively represent the hierarchical and parallel relationships in visual relationships. This tree structure provides a new research direction for scene graph representation. Fig. \ref{fig:Chain_Graph_VCTree} presents a comparison of the chain structure, fully connected graph structure, subgraph, and dynamic tree structure of the scene graph. 

SIG \cite{wang2020sketching} also proposed a scene graph with a similar tree structure; the key difference stems from the observation that humans tend to describe the subjects and key relationships in the image first when analyzing scenes, meaning that a hierarchy analysis with primary and secondary order is more in line with human habits. To this end, SIG proposed a human-mimetic hierarchical SGG method. Under this approach, the scene is represented by a human-mimetic HET (Hierarchical Entity Tree) composed of a series of image regions, while Hybrid-LSTM (Hybrid Long Short-Term Memory) is used to parse HET, thereby enabling the hierarchical structure \cite{tang2019learning} and siblings context \cite{zellers2018neural} information in HET to be obtained.

\vspace{-1em}
\subsection{GNN-based SGG}
\label{sec:Graph-based SGG}
\begin{figure} 
\center{\includegraphics[width=0.49\textwidth] {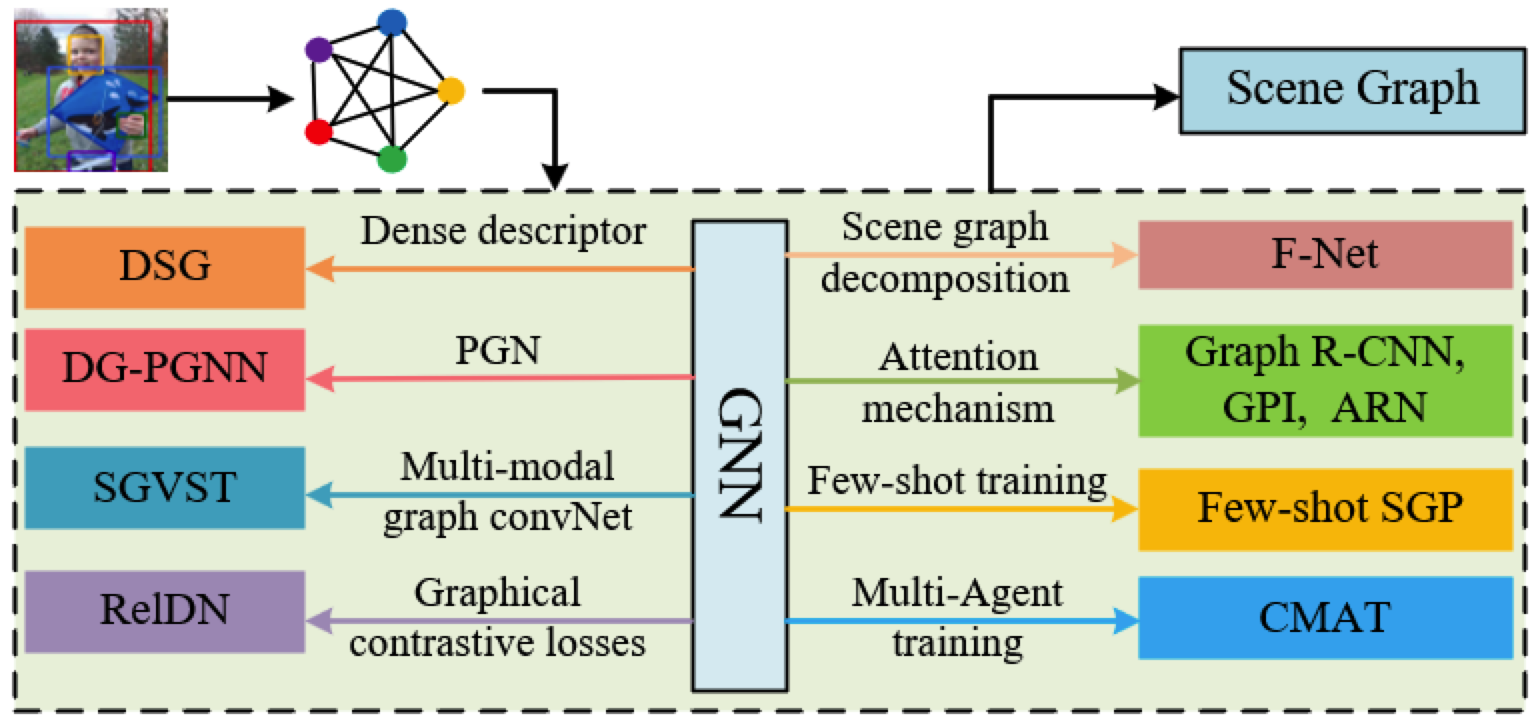}} 
\vspace{-2 em}
   \caption{Improved SGG models based on GNN. }
\vspace{-1.5 em}
 \label{graph}
\end{figure}

The scene graph can be regarded as a graph structure. An intuitive approach would therefore be to improve the generation of scene graphs with the help of graph theory. The GCN (Graph Convolutional Network) \cite{goller1996learning, gori2005new, scarselli2008graph, kipf2016semi} is just such a method. This approach is designed to process graph structure data, local information \cite{gilmer2017neural, Li2015Gated} can be effectively learned between neighboring nodes. GCN has been proven to be highly effective in tasks such as relational reasoning \cite{santoro2017simple}, graph classification \cite{bruna2013invariant, dai2016discriminative, defferrard2016convolutional, niepert2016learning}, node classification in large graphs \cite{hamilton2017inductive, kipf2016semi}, and visual understanding \cite{herzig2018classifying, herzig2018mapping, wang2018videos}. Accordingly, many researchers have directly studied the SGG method based on GCN. \textcolor{black}{ Similar to the conditional form (Eq.(\ref{eq:bottom_up})) of RNN/LSTM-based SGG methods, and following the expression form of the relevant variables in this paper, the SGG process based on GCN can also be factorized into three parts:}
\begin{equation}
P(\left \langle V,E,O,R \right \rangle|I) =P(V |I)*P(E|V, I)*P(R, O|V, E, I)
\label{eq:GCN-based}
\end{equation}
\noindent \textcolor{black}{ where $V$ is the set of nodes (objects in images), $E$ is the edges (relationships between objects) in a graph. Based on this basic conditional form, the subsequent improved GNN-based SGG methods are presented. Most of these methods attempt to optimize the terms of $P(E|V,I)$ and $P(R, O|V, E, I)$ by designing relevant modules, and GCN-based networks are designed for the graph labeling process $P(R, O|V, E, I)$. Fig. \ref{graph} presents some classic GNN-based SGG models.} F-Net (Factorizable Net) \cite{li2018factorizable} completes final SGG by decomposing and merging the graphs, after which the attention mechanism is introduced to design different types of GNN modules for SGG, such as Graph R-CNN \cite{yang2018graph}, GPI \cite{herzig2018mapping} and ARN (Attentive Relational Network) \cite{qi2019attentive}. Few-shot training and multi-agent training are applied to few-shot SGP \cite{dornadula2019visual} and CMAT (Counterfactual critic Multi-Agent Training) \cite{chen2019counterfactual} respectively. Probabilistic Graph Network (PGN) is designed for DG-PGNN \cite{khademi2020deep}, while multi-modal graph convNet was developed for SGVST \cite{wang2020storytelling}. Furthermore, other improved GNN-based network modules have been proposed for SGG, which we will describe in detail.

As discussed in Section \ref{sec:Construction process}, the current SGG methods can be roughly divided into two categories: \textit{bottom-up} and \textit{top-down} methods. However, these types of frameworks build a quadratic number of objects, which is time-consuming. Therefore, an efficient subgraph-based framework for SGG, called Factorizable Net (F-Net) \cite{li2018factorizable}, is proposed to promote the generation efficiency of the scene graph. With this approach, the detected object region proposals are paired to facilitate the construction of a complete directed graph. Thereafter, a more precise graph is generated by merging edges corresponding to similar union regions into a subgraph; each subgraph has several objects and their relationships are represented as edges. By substituting the original scene graph with these subgraphs, Factorizable Net can achieve higher generation efficiency of the scene graph.
Furthermore, Graph R-CNN \cite{yang2018graph} attempts to trim the original scene graph (removing those unlikely relationships) so as to generate a sparse candidate graph structure. Finally, an attention graph convolutional network (AGCN) is used to integrate global context information to promote more efficient and accurate SGG.

The graph-based attention mechanism also has important research value in the generation of scene graphs. For example, previous SGG work \cite{li2017scene, newell2017pixels, yang2018graph, herzig2018mapping, xu2017scene} often requires prior knowledge of the graph structure. In addition, these methods tend to ignore the overall structure and information of the entire image, as they capture the representation of nodes and edges in a step-by-step manner. Moreover, the one-by-one detection of the visual relationship of the paired regions \cite{jae2018tensorize, krishna2018referring, li2017vip, peyre2017weakly, yang2018shuffle, yin2018zoom} is also poorly suited to describing the structure of the entire scene.
For this reason, in ARN \cite{qi2019attentive}, a semantic transformation module is proposed that produces semantic embeddings by transforming label embeddings and visual features into the same space, while a relation inference module is used to predict the entity category and relationship as the final scene graph result. In particular, to facilitate describing the structure of the entire scene, ARN proposed a graph self-attention-based model aimed at embedding a joint graph representation to describe all relationships. This module helps to generate more accurate scene graphs. \textcolor{black}{In addition, an intuition is that when recognizing an image of a "person riding a horse", the interaction between the human leg and the horseback can provide strong visual evidence for the recognition of the predicate. For this reason, RAAL-SGG (Toward Region-Aware Attention Learning for SGG) \cite{liu2021toward} points out that it is limited to use only coarse-grained bounding boxes to study SGG. Therefore, RAAL-SGG proposed a region-aware attention learning method that uses an object-wise attention graph neural network for more fine-grained object region reasoning. The probability function of this model can be expressed as}
\begin{equation}
\begin{split}
P(SG|I)=&P(B|I)*P(F|B,I)*P(O|B,I,F)*\\&P(R|B,I,F,O),F=\{f^{(n)}\}_{n=1}^N,
\end{split}
\label{eq:RAAL-SGG}
\end{equation}
\textcolor{black}{where $f^{(n)}$ is a region set of the \textit{n}th object. Different from Eq.(\ref{eq:bottom_up}), Eq.(\ref{eq:RAAL-SGG}) considers the object area set $F$ which is more fine-grained than the coarse-grained bounding box $B$. This helps the model to reason about the predicate with the help of the object interaction area.}

When predicting the visual relationship of the scene graph, the reading order of entities in the context encoded using RNN/LSTM \cite{lee2019learning} also has a crucial influence on the SGG. A fixed reading order may not be optimal under these circumstances. A scene graph generator should reveal the connection between objects and relations to improve the prediction precision, even if different types of inputs are present. Formally, given the same features,  the same result should be obtained by a framework or a function $\mathcal{F}$ even if the input has been permuted. Motivated by this observation, the architecture of a neural network for SGG should ideally remain invariant to a particular type of input permutation. Herzig et al. \cite{herzig2018mapping} accordingly proved this property based on the fact that such an architecture or framework can gather information from the holistic graph in a permutation-invariant manner. Based on this feature, these authors proposed several common architecture structures and obtained competitive performance.

For most SGG approaches \cite{dai2017detecting, liao2019Natural, lu2016visual, yu2017visual, li2017vip, xu2017scene, yan2020pcpl}, the long-tailed distribution of relationships remains a challenge to relational feature learning. Existing methods are often unable to deal with unevenly distributed predicates. Therefore, Dornadula et al. \cite{dornadula2019visual} attempted to construct a scene graph via few-shot learning of predicates, which can scale to new predicates. The SGG model based on few-shot learning attempts to fully train the graph convolution model and the spatial and semantic shift functions on relationships with abundant data. For their part, the new shift functions are fine-tuned to new, rare relationships of a few examples. When compared to conventional SGG methods, the novelty of this model is that predicates are defined as functions, such that object notations are useful for few-shot predicate forecasting; these include a forward function that turns subject notations into objects and a corresponding function that changes the object representation back into subjects. The model achieves good performance in the learning of rare predicates.

A comprehensive, accurate, and coherent scene graph is what we expect to achieve, and the semantics of the same node in different visual relationships should also be consistent. However, the currently widely used supervised learning paradigm based on cross-entropy may not guarantee the consistency of this visual context. This is because this paradigm tends to predict pairwise relationships in an independent way \cite{lu2016visual, yang2018shuffle, zhang2017visual, shang2017video}, while hub nodes (those that belong to multiple visual relationships at the same time) and non-hub nodes are given the same penalty.  This is unreasonable. For this reason, \cite{chen2019counterfactual} proposed a Counterfactual critic Multi-Agent Training (CMAT) approach. More specifically, CMAT is the first work to define SGG as a cooperative multi-agent problem. This approach solves the problems of graph consistency and graph-level local sensitivity by directly maximizing a graph-level metric as a reward (corresponding to the hub and non-hub nodes being given different penalties).
Similarly, RelDN (Relationship Detection Network) \cite{zhang2019graphical} also found that applying cross-entropy loss alone may have an adverse effect on predicate classification; examples include Entity Instance Confusion (confusion between different instances of the same type) and Proximal Relationship Ambiguity (subject-object pairing problems in different triples with the same predicate).
RelDN is proposed to tackle these two problems. In RelDN, three types of features for semantic, visual, and spatial relationship proposals are combined by means of entity-wise addition. These features are then applied to obtain a distribution of predicate labels via softmax normalization. Thereafter, contrastive losses between graphs are specifically constructed to solve the aforementioned problems. 

Scene graphs provide a natural representation for reasoning tasks. Unfortunately, given their non-differentiable representations, it is difficult to use scene graphs directly as intermediate components of visual reasoning tasks. Therefore, DSG (Differentiable Scene-Graphs) \cite{raboh2020differentiable} are employed to solve the above obstacles. The visual features of objects are used as inputs to the differentiable scene-graph generator module of DSGs, which is a set of the new node and edge features. The novelty of the DSG architecture lies in its decomposition of the scene graph components, enabling each element in a triplet to be represented by a dense descriptor. Thus, DSGs can be directly used as the intermediate representation of downstream inference tasks.

Although we have investigated many GNN-based SGG methods, there are still many other related methods. For example, \cite{khademi2020deep} proposes a deep generative probabilistic graph neural network (DG-PGNN) to generate a scene graph with uncertainty. SGVST \cite{wangstorytelling} introduces a scene graph-based method to generate story statements from image streams. This approach uses GCN to capture the local fine-grained region representation of objects in the scene graph. We can conclude from the above that the GNN-based SGG method has attracted significant research attention due to its obvious ability to capture structured information.

\vspace{-1em}
\subsection{Discussion}
\textcolor{black}{As mentioned previously, ``the recognition of predicates is interrelated, and contextual information plays a vital role in the generation of scene graphs.''} To this end, researchers are increasingly focusing on methods based on RNN/LSTM or graphs. This is primarily because RNN/LSTM has better relational context modeling capabilities. While the graph structure properties of the scene graph itself also enable the GNN-based SGG to obtain the corresponding attention. In addition, TransE-based SGG has been welcomed by researchers due to its intuitive modeling method, which makes the model very interpretable. Due to the strong visual feature learning ability of CNN, future CNN-based SGG methods will remain mainstream SGG methods.

From the perspective of objects, relationships, and messages, the above methods can also be divided into object contextualization methods, relationship representation methods, message-passing methods, and relationship context methods. In the scene graph parsing task, all objects and relationships are related; accordingly, the corresponding context information should be fully considered. Most of the bounding box-based SGG methods consider the context information of the object \cite{newell2017pixels, yu2017visual, yin2018zoom, zhang2018interpretable}. \textcolor{black}{Because objects in these methods can be adequately interacting in the process of contexualizaiton of Link $I\rightarrow \widetilde{Y}$, Link $I\rightarrow X$ and subsequent $X\rightarrow \widetilde{Y}$.} Furthermore, due to the good modeling ability of RNN/LSTM in a relational context, the related SGG method is also favored by many researchers (as discussed in Section.\ref{sec:RNN/LSTM-based SGG}). There is a large amount of related work contained herein, including RNN \cite{chen2019panet, gao2019hierarchical}, GRU \cite{xu2017scene}, LSTM \cite{zellers2018neural, wang2020sketching}, TreeLSTM \cite{tang2019learning}. \textcolor{black}{Specifically, for example, in feature extraction (Link $I\rightarrow X$), in order to encode the visual context for each object, Motifs \cite{zellers2018neural} chooses to use Bi-LSTMs (bidirectional LSTMs). In PANet \cite{chen2019panet}, it uses the combination of class embedding, spatial information, and object visual features as the input of RNN in the contextualization process to obtain instance-level and scene-level context information.} The relationship representation method is also the main research direction at present because it directly affects the accuracy, completeness, and hierarchy of the relationship modeling. It mainly includes chains \cite{zellers2018neural}, fully connected graph \cite{xu2017scene, chen2018scene, chen2018iterative, dai2017detecting, li2018factorizable, xia2016weakly, yin2018zoom}, subgraph \cite{li2018factorizable}, and tree \cite{tang2019learning} structures (see Fig. \ref{fig:Chain_Graph_VCTree}). The related research helps in establishing a better explanatory model. \textcolor{black}{For example, MSDN (Multi-level Scene Description Network) \cite{li2017scene} builds a dynamic graph from three different levels of object, phrase, and region features in the process of contexualizaiton of Link $I\rightarrow X$, Link $X\rightarrow \widetilde{Y}$, and Link $I\rightarrow \widetilde{Y}$, in which the feature refining phase message can be passed on the edges of the graph. VCTree \cite{tang2019learning} uses Bi-TreeLSTMs (bidirectional TreeLSTMs) \cite{tai2015improved} to encode visual context in feature extraction (Link $I\rightarrow X$).} In addition, message passing \cite{xu2017scene, dhingra2021bgt} is also an important research direction, as it directly affects the degree of information interaction between objects and relationship nodes (as discussed in Section.\ref{sec:Graph-based SGG}). In the future, the mining of relevant contextual information will continue to be a highly promising research direction. In addition, it is also necessary to add reasoning capabilities to the model, as this will aid in solving the long-tail problem.

\textcolor{black}{In addition to the above-discussed method of using full supervision for SGG, \cite{ye2021linguistic} proposes a new method of using weak supervision to generate scene graphs. It uses the image and the corresponding caption as weak supervision to learn the entities in the image and the relationship between them. This further expands the available dataset for scene graph related research. 
In addition, PUM (Probabilistic Uncertainty Modeling) \cite{yang2021probabilistic} proposes that in order to generate an ‘accurate’ scene graph, it is inappropriate to predict the relationship between entities in a deterministic way. This is mainly due to the semantic ambiguity (e.g., synonymy ambiguity, hyponymy ambiguity, and multi-view ambiguity) of visual relationship. For this reason, PUM uses probabilistic uncertainty modeling to replace the previous deterministic relationship with predictive modeling in the construction and inference stage of the graph. This strategy enables PUM to generate more realistic and diversified scene graphs.
Similar multi-directional research is necessary for the development of scene graphs.}

\vspace{-1em}
\section{SGG with prior knowledge}
\label{sec:SGG_Proir}
For SGG, relationships are combinations of objects, and its semantic space is wider than that of the objects. In addition, it is very difficult to exhaust all relationships from the SGG training data. It is therefore particularly critical to effectively learn relationship representations from a small amount of training data. The introduction of prior knowledge may thus greatly assist in the detection and recognition of visual relationships. Therefore, in order to efficiently and accurately generate a complete scene graph, the introduction of prior knowledge (such as language prior, visual prior, knowledge prior, context, etc.) is also crucial.  In this section, we will introduce the related work of SGG with prior knowledge.

\vspace{-1em}
\subsection{SGG with Language Prior}
\begin{figure} 
\center{\includegraphics[width=0.48\textwidth] {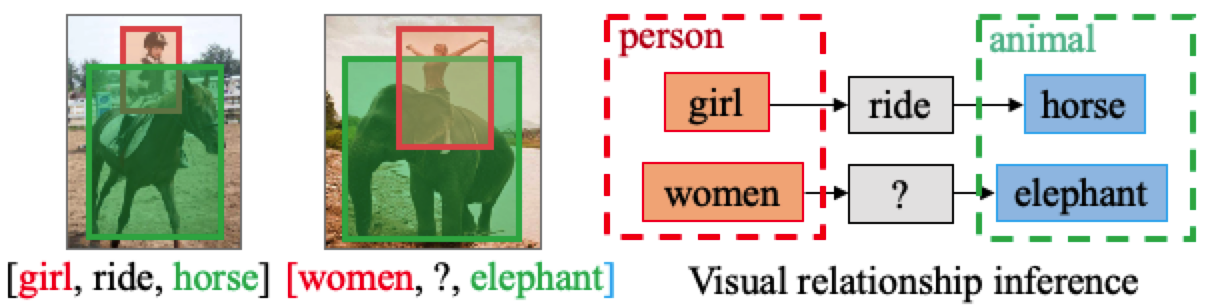}}
\vspace{-1 em}
   \caption{Examples of semantic similarity helping with visual relationship inference. 
   }
   \vspace{-1.5 em}
 \label{fig:girl_women_ride}
\end{figure}

Language priors typically use the information embedded in semantic words to fine-tune the possibility of relationship prediction, thereby improving the accuracy of visual relationship prediction. Language priors can help the recognition of visual relationships through the observation of semantically related objects. For example, horses and elephants may be arranged in a semantically similar context, e.g., "a person riding a horse" and "a person riding an elephant". Therefore, although the co-occurrence of elephants and persons is not common in the training set, through the introduction of language priors and the study of more common examples (such as "a person riding a horse"), we can still easily infer that the relationship between a person and an elephant may be one of riding. This idea is illustrated in Fig.\ref{fig:girl_women_ride}. This approach also helps to resolve the long tail effect in visual relationships.

Many researchers have conducted detailed studies on the introduction of language priors. For example, Lu et al.\cite{lu2016visual} suggested training a visual appearance module and a language module simultaneously, then combining the two scores to infer the visual relationship in the image. In particular, the language a priori module projects the semantic-like relationships into a tighter embedding space.  This helps the model to infer a similar visual relationship ("person riding an elephant") from the "person riding a horse" example. Similarly, VRL (deep Variation-structured Reinforcement Learning) \cite{liang2017deep} and CDDN (Context-Dependent Diffusion Network) \cite{cui2018context} also use language priors to improve the prediction of visual relationships; the difference is that \cite{lu2016visual} uses semantic word embedding \cite{mikolov2013efficient} to fine-tune the possibility of predicting relationships, while VRL follows the variational-structured traversal scheme over a directed semantic action graph from the language prior, meaning that the latter can provide richer and more compact semantic association representation than word embedding. 
Moreover, CDDN finds that similar objects have close internal correlations, which can be used to infer new visual relationships. To this end, CDDN uses word embedding to obtain a semantic graph, while simultaneously constructing a spatial scene graph to encode global context interdependency. CDDN can effectively learn the latent representation of visual relations through the combination of prior semantics and visual scenes; furthermore, considering its isomorphic invariance to graphs, it can cater well to visual relation detection.

On the other hand, although the language prior can compensate for the difference between model complexity and dataset complexity, its effectiveness will also be affected when the semantic word embedding falls short \cite{atzmon2016learning}. For this reason, \cite{jae2018tensorize} further introduces a relation learning module with a priori predicate distribution on the basis of IMP \cite{xu2017scene} to better learn visual relations. 
In more detail, a pre-trained tensor-based relation module is added to \cite{jae2018tensorize} as a dense relation prior to fine-tuned relation estimation, while an iterative message-passing scheme with GRUs is used as a GCN method for promoting the SGG performance with better feature representation. 
In addition to using language priors, \cite{plummer2017phrase} also combines visual cues to identify visual relationships in images and locate phrases. For its part, \cite{plummer2017phrase} models the appearance, size, location, and attributes of entities, along with the spatial relationship between object pairs connected by verbs or prepositions, and jointly infers visual relationships through automatically learning and combining the weights of these clues.

\vspace{-1em}
\subsection{SGG with Statistical Prior}
Statistical prior is also a form of prior knowledge widely used by SGG, as objects in the visual scene usually have strong structural regularity \cite{zellers2018neural}. For example, people tend to wear shoes, while mountains tend to have water around them. In addition, $\langle cat-eat-fish \rangle$ is common, while $\langle fish-eat-cat \rangle$ and $\langle cat-ride-fish \rangle$ are very unlikely. This relationship can thus be expressed using prior knowledge of statistical correlation. Modeling the statistical correlation between object pairs and relationships can help us in correctly identifying visual relationships.

Due to the spatially large and long-tailed nature of relationship distributions, simply using the annotations contained in the training set would be insufficient.  Moreover, it is difficult to collect an adequate amount of labeled training data. For this reason, LKD (Linguistic Knowledge Distillation) \cite{yu2017visual} uses not only the annotations inside the training set but also text publicly available on the Internet (Wikipedia) to collect external language knowledge. This is mainly achieved by tallying the vocabulary and expressions used by humans to describe the relationships between pairs of objects in the text, then calculating the conditional probability distribution ($P(pred|subj, obj)$) of the predicate given a pair of $\langle subj, obj \rangle$. A novel contribution is that of using knowledge distillation \cite{Hu2016Harnessing} to acquire prior knowledge from internal and external linguistic data in order to solve the long-tail relationship problem.

Similarly, DR-Net (Deep Relational Networks) \cite{dai2017detecting} also noticed the strong statistical correlation between the triples $\langle subj-pred-obj \rangle$.  The difference is that DR-Net proposed a deep relationship network to take advantage of this statistical correlation. DR-Net first extracts the local regions and spatial masks of each pair of objects, then inputs them together with the appearance of a single object into the deep relational network for joint analysis, thereby obtaining the most likely relational category.
In addition, MotifNet \cite{zellers2018neural} performed a statistical analysis of the co-occurrences between the relationships and object pairs on the Visual Genome dataset \cite{krishna2017visual}, finding that these statistical co-occurrences can provide strong regularization for relationship prediction. To this end, MotifNet uses LSTM \cite{hochreiter1997long} to encode the global context of objects and relationships, thus enabling the scene graph to be parsed. 
However, although the above methods \cite{zellers2018neural, dai2017detecting} also observed the statistical co-occurrence of the triple, the depth model they designed implicitly mined this statistical information through message transmission.
KERN (Knowledge-Embedded Routing Network) \cite{chen2019knowledge} also took note of this statistical co-occurrence. The difference is that KERN formally expresses this statistical knowledge in the form of a structured graph, which is incorporated into the deep propagation network as additional guidance. This can effectively regularize the distribution of possible relationships, thereby reducing the ambiguity of prediction.

In addition, similar statistical priors are also used in complex indoor scene analysis \cite{yang2017support}.  Statistical priors can effectively improve performance on the corresponding scene analysis tasks.

\vspace{-1em}
\subsection{SGG with Knowledge Graph}
Knowledge graphs are a rich knowledge base that encodes how the world is structured. Common-sense knowledge graphs have thus been used as prior knowledge to effectively help the generation of scene graphs.

To this end, GB-Net (Graph Bridging Network) \cite{zareian2020bridging} proposes a new perspective, which constructs scene graphs and knowledge graphs into a unified framework. More specifically, GB-Net regards the scene graph as the image-conditioned instantiation of the commonsense knowledge graph. Based on this perspective, the generation of scene graphs is redefined as a bridge mapping between scene graphs and common sense graphs. 
In addition, the deviations in the existing label dataset on object pairs and relationship labels, along with the noise and missing annotations they contain, all increase the difficulty of developing a reliable scene graph prediction model. For this reason, KB-GAN (knowledge base and auxiliary image generation) \cite{gu2019scene} proposed a SGG algorithm based on external knowledge and image reconstruction loss to overcome the problems found in datasets. More specifically, KB-GAN uses ConceptNet's \cite{speer2013conceptnet} English subgraph as the knowledge graph; the knowledge-based module of KB-GAN improves the feature refinement process by reasoning on a basket of common sense knowledge retrieved from ConceptNet. Similarly, there are many related works \cite{chen2019knowledge, herzig2018mapping, khademi2020deep, Sharifzadeh2021ClassificationBA} that use knowledge graphs as prior knowledge to assist relationship prediction.

\vspace{-1em}
\subsection{Discussion}
\begin{figure} 
\center{\includegraphics[width=0.49\textwidth] {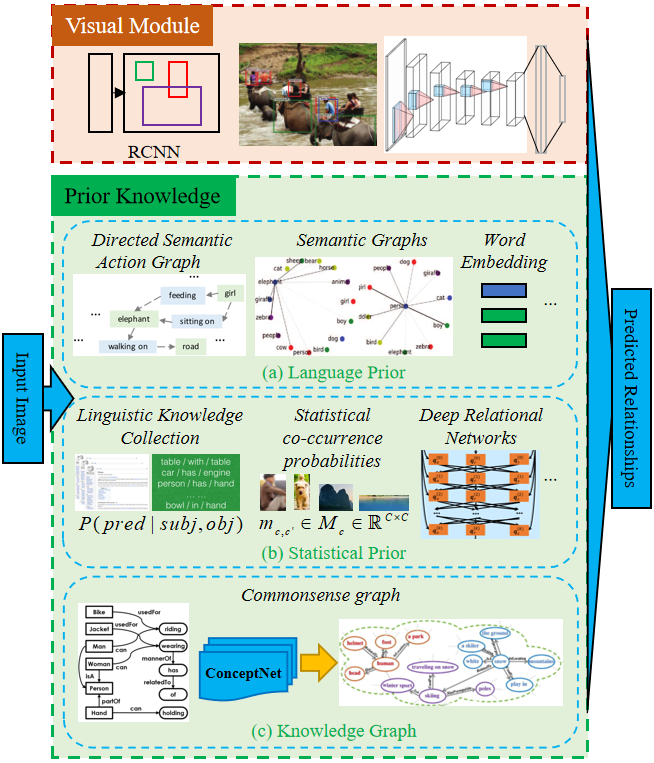}} 
\vspace{-1.5 em}
   \caption{A schematic diagram of a SGG framework assisted by prior knowledge. Such models often consist of two branches: a visual module and an additional prior knowledge module. 
Prior knowledge includes (a) language priors (such as directed semantic action graphs \cite{liang2017deep}, semantic graphs \cite{cui2018context}, or word embedding \cite{cui2018context, lu2016visual}), (b) statistical priors (such as linguistic knowledge 
collection \cite{yu2017visual}, statistical co-occurrence probabilities \cite{chen2019knowledge} and Deep Relational Networks \cite{dai2017detecting}) and (c) knowledge graphs \cite{zareian2020bridging, gu2019scene}. 
}
\vspace{-1.5 em}
 \label{fig:priors}
\end{figure}

Fig.\ref{fig:priors} presents the pipeline of SGG models with different types of prior knowledge. 
Prior knowledge has been proven to significantly improve the quality of SGG. Existing methods use either an external curated knowledge base, such as ConceptNet \cite{gu2019scene, lee2018multi, wang2018zero, speer2017conceptnet}, or the statistical information found in the annotation corpus to obtain commonsense data. However, the former is limited by incomplete external knowledge \cite{chen2019knowledge, yu2017visual, su2018learning, zellers2018neural}, while the latter is often based on hard-coded heuristic algorithms such as the co-occurrence probability of a given category. Therefore, the latest research \cite{zareian2020learning} attempts to use visual commonsense as a machine learning task for the first time, and automatically obtains visual commonsense data directly from the dataset to improve the robustness of scene understanding. While this exploration is very valuable, the question of how to acquire and make full use of this prior knowledge remains a difficult one that merits further attention.

\vspace{-1em}
\section{Long-tailed distribution on SGG}
\label{sec:long-tail}
\begin{figure} 
\center{\includegraphics[width=0.4\textwidth] {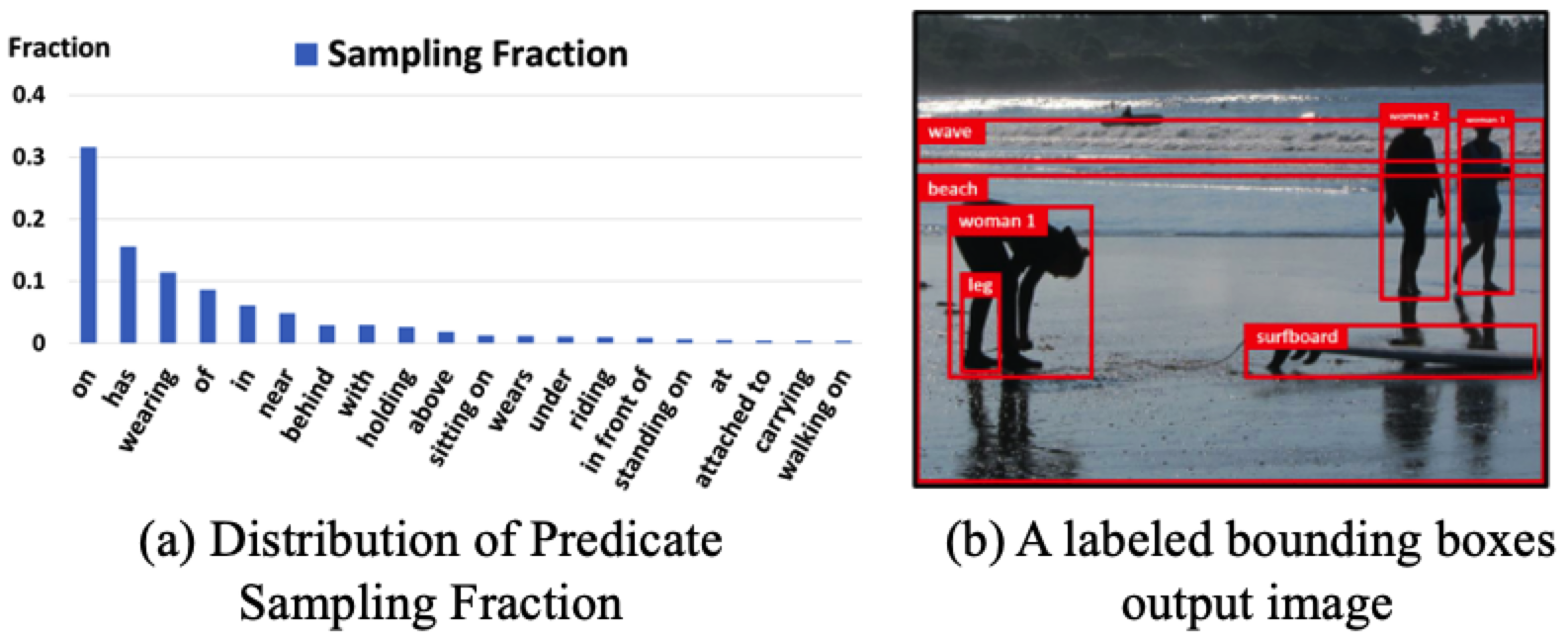}}  
\vspace{-1 em}
   \caption{An example of SGG. (a) The distribution of the 20 most common predicates in Visual Genome \cite{ji2019action}. (b) A labeled bounding boxes output image. 
   Copied from \cite{Tang2020}.
}
\vspace{-1.5 em}
 \label{fig:Long-tail}
\end{figure}

Long-tail distribution is a key challenge in visual relationship recognition. Predicates in visual relationships are often unevenly distributed, and common head predicates (such as ``on'', ``have'', ``in'', etc.) occupy most of the relationship expression (as shown in Fig. \ref{fig:Long-tail} (a)). However, these overly general relationship expressions have very limited significance for the analysis of visual relationships. We would prefer to see more accurate expressions, such as ``people walking on the beach'' or ``people lying on the beach'' rather than ``people on the beach'' (as shown in Fig. \ref{fig:Long-tail} (b)). 
When analyzing the complexity of visual relationships, it should be noted that because visual relationships combine objects and predicates, the complexity of this relationship is $\mathcal{O}(N^2R)$ for $N$ objects and $R$ predicates. Even if learning the models for the object and predicate separately reduces the complexity to $\mathcal{O}(N+R)$, the dramatic changes in the appearance of the predicate remain very challenging. For example, there is a significant difference in the visual appearance of $\langle person-ride-bike \rangle$ and $\langle person-ride-horse \rangle$ (as shown in Fig. \ref{fig:unsee_relation}). The distribution of such object-predicate combinations tends to be more long-tailed than that of objects alone. 
In addition, due to the inherent complexity of real scenes, the fact that some visual relationships have never appeared in the training set does not mean that they do not exist, and it is impossible to exhaust them in the training set (e.g., the right side of Fig. \ref{fig:unsee_relation}). Therefore, the study of the long-tail problem is urgent and necessary.

In scene graph datasets, the actually existing long-tailed distribution of relationships directly affects the accuracy and completeness of the generated scene graph and is thus a problem that many scholars have been trying to solve for the SGG context. For example, zero-shot \cite{lee2018multi, shen2018scaling, gkanatsiossaturation}, one-shot \cite{guo2020one} and few-shot \cite{dornadula2019visual, knyazev2020graph} learning approaches attempt to address the challenges of scaling relationship recognition to the long tail of categories in the datasets. Specifically, \cite{dornadula2019visual} learns a meaningful embedding space by treating the predicate as a function between object representations so that the predicate can be transferred between objects of the same kind, thereby enhancing the learning ability of the model in the few-shot problem. 
In addition, \cite{knyazev2020graph} proposes that the standard scene graph density loss function can result in individual edges in a large sparse graph being readily ignored during the training process, and the ``frequent'' relationship prediction leads to a ``blind'' model that can also achieve good performance. For this reason, \cite{knyazev2020graph} introduced a density-normalized edge loss and proposed a novel weighted metric for zero/few shots, which achieved good results.  Moreover, the language prior information \cite{liang2017deep, cui2018context,lu2016visual} and statistical prior  \cite{yu2017visual, dai2017detecting, zellers2018neural, chen2019knowledge} are used to project relationships, enabling similar, rare relationships to be predicted so as to alleviate the problem of the long tail of infrequent relationships. Such types of prior information or analogies between similar relationships are very helpful for detecting infrequent relationships. 

In addition, \cite{he2020learning} aims to solve the long-tail distribution problem in the scene graph by transferring the knowledge learned from the head relationship (relationship with a larger order of magnitude instances) to the tail (relationship with a smaller order of magnitude instance) by means of knowledge transfer. 
The deviation caused by the long-tail problem is widespread. Traditional debiasing methods are often unable to distinguish between good deviation (e.g., people riding a bicycle instead of eating) and bad deviation (e.g., on instead of walking/lying on). For this reason, TDE (Total Direct Effect) \cite{Tang2020} explored a method based on causal inference biased training. TDE finds and eliminates bad bias by extracting counterfactual causality from the training graph.
\textcolor{black}{Recently, EBM (Energy Based Modeling) \cite{suhail2021energy} pointed out that those methods \cite{tang2019learning, zellers2018neural, xu2017scene, vaswani2017attention} that use standard cross-entropy loss still essentially treat the prediction of objects and relationships in the scene graph as independent entities.
\begin{equation}
    log P(SG|I)=\sum_{o_i\in O}log P(o_i,o_i^{bbox}|I)+\sum_{r_{i\rightarrow j\in R}}log P(r_{i\rightarrow j}|I).
    \label{eq:EBM}
\end{equation}
The modification of Eq.(\ref{eq:EBM_0}), Eq(\ref{eq:EBM}), explains the nature of this independent prediction. To this end, EBM proposes to use an energy-based learning framework \cite{lecun2006tutorial} to generate scene graphs. This framework allows the use of "loss" that explicitly incorporates structure in the output space to train the SGG model. It still achieved good results under zero- and few-shot settings.} Similarly, exploring in different directions is very beneficial to alleviating the long-tail problem.

Although researchers have done a lot of related research, the long-tail problem will remain a continuing hot issue in scene graph research.

\vspace{-1em}
\section{Applications of Scene Graph}
\label{sec:SGG_Applications}
\begin{figure*} 
	\centering
	\includegraphics[width=0.9\linewidth]{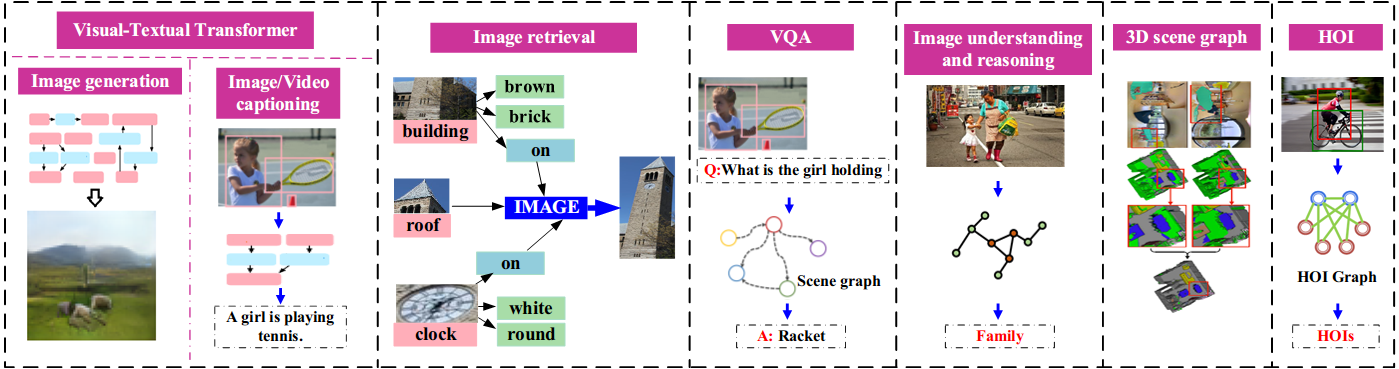}
	\vspace{-1 em}
	\caption{Examples of scene graph application scenarios. These applications include visual-textual transformers \cite{johnson2018image, Mittal2019Interactive, tripathi2019using}, image text retrieval \cite{johnson2015image, schuster2015generating, qi2017online}, visual question answering \cite{yang2018scene, ghosh2019generating}, image understanding and reasoning \cite{aditya2018image,shi2019explainable,zhang2019large}, 3D scene graphs \cite{armeni20193d,kim20203d}, and the detection and recognition of human-object and human-human interaction \cite{li2018transferable,zhuang2018hcvrd,shen2018scaling,chao2018learning,gkioxari2018detecting}.}
	\vspace{-1.5 em}
	\label{fig:SGG_Applications}
\end{figure*}
The scene graph can describe the objects in a scene and the relationships between the objects, meaning that it provides a better representation for scene understanding-related visual and textual tasks and can greatly improve the model performance of these tasks. 
Fig.\ref{fig:SGG_Applications} presents some examples of scene graph application scenarios. Next, we will conduct a detailed review of these scene graph applications one by one. 

\vspace{-1em}
\subsection{Visual-Textual Transformer}

Scene graphs contain the structured semantic information in a visual scene, and this semantic information is mainly reflected in the representations of objects, attributes, and pairwise relationships in images. Thus, scene graphs can provide favorable reasoning information for the vision-text tasks of image generation and image/video captioning.

\vspace{-1em}
\subsubsection{Image Generation}

Although text-based image generation has made exciting progress in the context of simple textual descriptions, it is still difficult to generate images based on complex textual descriptions containing multiple objects and their relationships. Image generation based on scene graphs is better able to deal with complex scenes containing multiple objects and desired layouts.

Generating complex images from layouts is more controllable and flexible than text-based image generation. However, it remains a difficult one-to-many problem, and only limited information can be conveyed by a bounding box and its corresponding label. To generate a realistic image according to the corresponding scene graph with object labels and their relationships, Johnson et al. \cite{johnson2018image} proposed an end-to-end image generation network model. Compared with text-based image generation methods, a final complex image with many recognizable objects can be generated using this method by explicitly inferring the objects and relationships based on structured scene graphs. However, this image generation method \cite{johnson2018image} cannot introduce new or additional information to existing descriptions and is limited to generating images at one time. Therefore, Mittal et al. proposed a recursive network architecture \cite{Mittal2019Interactive} that preserves the image content generated in previous steps and modifies the accumulated images based on newly provided scene information. This method allows the context of sequentially generated images to be preserved by subjecting certain information to subsequent image generation conditions. However, there are still problems associated with ensuring that the generated image conforms to the scene graph and measuring the performance of the image generation models. Subsequently, an image generation method was proposed that harnesses the scene graph context and improves image generation \cite{tripathi2019using}. The scene context network encourages the generated images not only to appear realistic but also to respect the scene graph relationships. Similarly, Layout2Im (Layout-based Image generation) \cite{zhao2019image} is also an end-to-end model for generating images from layouts. Different from other related methods, Layout2Im breaks down the representations of each object into specified and unspecified parts, and individual object representations are grouped together for encoding the layouts.

From \cite{johnson2018image,Mittal2019Interactive,tripathi2019using,zhao2019image}, we can see that generating a layout from a scene graph is an important step in layout-based image generation. Therefore, Herzig et al.\cite{herzig2020learning} and Tripathi et al. \cite{tripathi2019compact} attempted to improve the quality of images generated from scene graphs by generating better layouts. Generating realistic images with complex visual scenes is a challenging task, especially when we want to control the layouts of the generated images. Herzig et al. \cite{herzig2020learning} present a novel SRC (Soft Relations Closure) Module to inherently learn the canonical graph representations, with the weighted graph representations obtained from a GCN used to generate the scene layouts. SRC can canonicalize graphs to improve layout generation. Moreover, Tripathi et al. \cite{tripathi2019compact} proposed a scene layout generation system that generates structured scene layouts. 
Similarly, to solve the layout prediction problem, Schroeder et al. proposed a layout prediction framework based on Triplet-Aware Scene Graph Embeddings \cite{schroeder2019triplet}. Triplet embeddings with supervisory signals are used to improve scene layout prediction, while a data augmentation technique is utilized to maximize triplet numbers during training. These two methods of additional supervision and data augmentation can enhance the embedding representation, enabling better layout outputs to be obtained.

The above-mentioned image generation method of the scene graph is based on the layout, and aims to ensure the semantic consistency between the generated image and the scene graph. However, they often overlook the visual details of the images and objects. PasteGAN \cite{yikang2019pastegan} is a semi-parametric method for image generation based on scene graphs and object cropping, designed to improve the visual quality of generated images. The proposed Crop Refining Network and Object-Image Fuser with attention mechanism in PasteGAN can encode objects' spatial arrangement, appearances, and interactions. Compared with most scene graph image generation methods, PasteGAN can parameterize control of the appearance of objects in the generated images. In addition, one interesting approach to image generation utilizes image collections to generate a narrative collage based on scene graphs \cite{Fang18}. In this process, the object relationship is crucial to the object positions in the narrative collage. However, the scene graphs used here to build and evaluate are primarily rule-based. 
Furthermore, SS-SGN (Spatio-Semantic Scene Graph Network) \cite{dhamo2020semantic} proposes a scene graph-based image modification method that can interact with users. More specifically, the user only needs to change a certain node or edge of the scene graph and then apply this change to edit the image. This provides users with more flexibility to modify or integrate new content into the image.

\vspace{-1em}
\subsubsection{Image/Video Captioning}
Traditional image captioning methods rely on the visual features of the objects detected in images. While, the natural language description is generated by models of natural language reasoning, such as RNN or LSTM. However, these methods cannot make full use of the semantic relationships between objects, which leads to the generated language description being inaccurate. The image captioning methods based on scene graphs solve this problem by capturing the relationship information between objects. 

The image captioning method in \cite{gao2018image} tried to learn the semantic representation by embedding a scene graph as an intermediate state. This method is easy to execute, does not require complex image preprocessing, and is competitive with existing methods. Since graphical representations with conceptual positional binding can improve image captioning, TPsgtR (Tensor Product Scene-Graph-Triplet Representation) \cite{sur2019tpsgtr} is proposed for image caption generation using regional visual features. In TPsgtR, the technique of neuro-symbolic embedding can embed the relationships into concrete forms (neural symbolic representations), rather than relying on the model to form all possible combinations. These neural symbolic representations aid in defining the neural symbolic space and can be transformed into better captions. In addition, the visual relational features can be learned from a neural scene graph generator (Stacked Motif Network) \cite{lee2019learning}, and will facilitates the grounding of language in visual relations.

From the perspective of human cognition, vision-based language generation is related to high-level abstract symbols. Abstracting the scenes into symbols will accordingly provide a clear path to language description generation. Therefore, SGAE (Scene Graph Auto-Encoder) \cite{yang2019auto} is proposed to incorporate these inductive biases into the encoder-decoder models for image captioning, an approach expected to help this encoder-decoder model exhibit less overfitting to the dataset bias. 
Similarly, to be able to generate the type of image descriptions desired by human, Chen et al. proposed an ASG (Abstract Scene Graph) structure \cite{chen2020say} to represent user intentions, as well as to control the generated descriptions and detailed description of the scenes. ASG can identify users' intention and semantics in graphs, which enables it to generate the required caption, and actively considers users' intentions to produce the desired image caption, which significantly enhances the image caption diversity. Unlike ASG, which generates diversified captions by focusing on different combinations of objects, SGD (Scene Graph Decomposition) \cite{zhong2020comprehensive} is to decompose the scene graph into a set of subgraphs, then use a deep model to select the most important subgraphs. SGD can obtain accurate, diverse, and controllable subtitles through the use of subgraphs.

In previous work, entities in images are often considered separately, which leads to the lack of structured information in the generated sentences. Scene graphs are structured by leveraging both visual features and semantic knowledge, and image captioning frameworks are proposed based on the structural-semantic information \cite{li2019know,xu2019scene,wangstorytelling}. In \cite{li2019know}, a hierarchical-attention-based module is designed to learn the correlation scores of the visual features and semantic relationship features, which are used to obtain the final context feature vector rather than simply concentrating the two feature vectors into a single vector. SGC (Scene Graph Captioner) \cite{xu2019scene} captures the integrated structural-semantic features from visual scenes, after which LSTM-based models translate these semantic features into the final text description. 
Furthermore, a scene graph can also be used to generate the story from an image stream; the SGVST proposed in \cite{wangstorytelling} can model the visual relations both within one image and across images, which is conducive to image captioning. This method significantly improves the fluency and richness of the generated stories.

At present, most image captioning models rely heavily on image-caption pair datasets, while unpaired image captioning presents great challenges when it comes to extracting and mapping different features of visual and textual modalities. Therefore, there are high costs associated with obtaining large-scale paired data of images and texts. To solve this problem, an unpaired scene graph-based image captioning approach is presented in \cite{gu2019unpaired} to capture rich semantic information from scenes. It further proposes an unsupervised feature extraction method to learn the scene graph features by mapping from the visual features of the image to the textual features of the sentences.

\vspace{-1em}
\subsection{Cross-Modal Retrieval}
Cross-modal retrieval is also a common application in scene graph research.
Image-text retrieval is a classic multi-modal retrieval task. The key to image-text cross-modal retrieval concerns learning a comprehensive and unified representation to represent multi-modal data. The scene graph is a good choice in this context. 

Image retrieval via scene graph was first applied in \cite{johnson2015image}. By replacing textual descriptions with scene graphs for image retrieval, the model can accurately describe the semantics of images without unstructured texts, and can further retrieve related images in more open and interpretable retrieval tasks. Subsequently, another scene graph-based image retrieval method was proposed by Schuster et al.\cite{schuster2015generating}. These may be the two earliest methods to involve the construction and applications of the scene graph. The results of their experiments show that image retrieval using scene graphs achieves better results than traditional image retrieval methods based on low-level visual features.
At present, most existing cross-modal scene retrieval methods ignore the semantic relationship between objects in the scene and the embedded spatial layout information. Moreover, these methods adopt a batch learning strategy and are thus unsuitable for processing stream data. To solve these problems, an online cross-modal scene retrieval method \cite{qi2017online} is proposed that utilizes binary representations and semantic graphs. 

The semantic graph can serve as a bridge between the scene graph and the corresponding text that enables the measurement of the semantic correlation between different modal data. 
However, most text-based image retrieval models experience difficulties when searching large-scale image data, such that the model needs to resort to an interactive retrieval process through multiple iterations of question-answering. To solve this problem, Ramnath et al. proposed an image retrieval framework based on scene graph \cite{ramnath2019scene}, which models the retrieval task as a learnable graph-matching problem between query graphs and catalog graphs. Their approach incorporated the strategies and structural constraints of the retrieval task into inference modules using multi-modal graph representation. 
Similarly, SQIR (Structured Query-based Image Retrieval using scene graphs) \cite{schroeder2020structured} is also a scene graph based image retrieval framework. The difference is that SQIR determined that structured queries (e.g. "little girl riding an elephant") are more likely to capture the interaction between objects than single-object queries (e.g. "little girl", "elephant"). To this end, SQIR proposes an image retrieval method based on scene graph embedding, which treats visual relationships as directed subgraphs of the scene graph for the purposes of the structured query.
In addition, Wang et al. \cite{wang2020cross} proposed two kinds of scene graphs (visual scene graph (VSG) and textual scene graph (TSG)) to represent image and text respectively, and matched two different modalities through the unified representation of scene graphs.
However, the above methods are often based on fixed text or images for cross-modal retrieval. GEDR (Graph Edit Distance Reward) \cite{chen2020graph} proposes a more creative and interactive image retrieval method. More specifically, similar to SS-SGN \cite{dhamo2020semantic}, GEDR attempts to edit the scene graph, while GEDR edits the scene graph according to the user's text instructions on the given image prediction scene graph to perform image retrieval. This makes image retrieval more flexible and promotes easier user interaction. 

In addition, the scene graph is also used as an intermediate representation of 2D-3D and 3D-3D matching \cite{wald2020learning}. Although there is not much related work, scene graphs are still expected to perform well in other similar cross-modal retrieval tasks.

\begin{table*}[t]
    \centering
    \caption{Statistics of the scene graph dataset. "-" indicates that this attribute is not released, \textcolor{black}{and "\XSolidBrush" indicates that the attribute is not applicable.}}
    \vspace{-1 em}
    \resizebox{\textwidth}{!}{
    \begin{tabular}{|c|l|l|c|cc|ccc|cc|}
        \hline
        &\multirow{3}{*}{Dataset} &\multirow{3}{*}{Size}&\makecell[c]{\#Action}& \multicolumn{2}{c|}{Objects}& \multicolumn{3}{c|}{Relationships} & \multicolumn{2}{c|}{Predicates}\\ 
        & && categories & \#bbox & \#categories & \#triplet & \#categories &\makecell[c]{\#zero-shot\\w \& w/o (categories)}&\#categories &  \makecell[c]{\#predicates per\\object category}\\ \hline
        \multirow{10}{*}{Image}&RW-SGD\cite{johnson2015image}&5K&\XSolidBrush&94K&7K & 113K&1K&-&-&3.3\\
        &VRD \cite{lu2016visual}&5K&\XSolidBrush&-&100&38K&7K&-&-&24.25\\
        &UnRel \cite{peyre2017weakly}&1K&\XSolidBrush&-&-&76&-&-&-&-\\
        &HCVRD \cite{zhuang2018hcvrd}&53K&\XSolidBrush&-&1.8K&257K&28K&18.5K \& 9.8k&927 &10.63\\
        &Open Images \cite{xu2017scene} & - &\XSolidBrush& 3M & 57 & 375K & 10 & - &- & -\\
        &Visual Phrase \cite{Sadeghi2011Recognition} &3K&\XSolidBrush&3K&8&1.8K & 17& -&- & 120\\
        &Scene Graph \cite{schroeder2020structured}& 5K &\XSolidBrush& 69K & 266 & 110K & 23K& -&- &2.3\\
        &VG \cite{krishna2017visual}&108K &\XSolidBrush &3.8M & 33.8K &-& 42K& -&- &-\\
        &VrR-VG \cite{liang2019vrr}&59K & \XSolidBrush & 282K & 1.6K & 203K & 117& -&- &-\\
        &VG150 \cite{xu2017scene}& 88K & \XSolidBrush & 739K & 150 & 413K & 50& -&- &-\\ \hline
        
        \multirow{2}{*}{Video}&CAD120++ \cite{zhuo2019explainable} & 0.5K (0.57\textit{h}) & 10 & 64K & 13 & 32K & 6& -&- &-\\
        &Action Genome \cite{ji2019action}&10K (82\textit{h}) & 157 & 0.4M & 35 & 1.7M & 25& -&- &-\\ \hline
        
        \multirow{4}{*}{3D}&Armeni et al. \cite{armeni20193d} & \makecell[l]{35 buildings\\727 rooms} & \XSolidBrush & 3K & 28 & - & 4& -&- &-\\
        &3DSSG \cite{wald2020learning}&\makecell[l]{1482 scans \\478 scenes} & \XSolidBrush & 48K & 534 & 544K & 40& -&- &-\\
        \hline
    \end{tabular}
    }
    \label{tab:datasets}
    \vspace{-1 em}
\end{table*}

\vspace{-1em}
\subsection{Visual Question Answering}
VQA is also a multimodal feature learning task. Compared with traditional VQA methods, scene graphs can capture the essential information of images in the form of graph structures, which helps scene graph-based VQA methods outperform traditional algorithms. 

Inspired by the application of traditional QA systems on knowledge graphs, an alternative scene graph-based approach was investigated \cite{zhang2019empirical}. Zhang et al. explored how to effectively use scene graphs derived from images for visual feature learning, and further applied graph networks (GN) to encode the scene graph and perform reasoning according to the questions provided. Moreover, Yang et al. \cite{yang2018scene} aimed to improve performance on VQA tasks through the use of scene graphs, and accordingly proposed a new model named Scene GCN (Scene Graph Convolutional Network) \cite{yang2018scene} to solve the relationship reasoning problem in a visual question-and-answer context. To effectively represent visual relational semantics, a visual relationship encoder is built to yield discriminative and type-aware visual relationship embeddings, constrained by both the visual context and language priors. To confirm the reliability of the results predicted by VQA models, Ghosh et al. \cite{ghosh2019generating} proposed an approach named XQA (eXplainable Question Answering), which may be the first VQA model to generate natural language explanations. In XQA, natural language explanations that comprise evidence are generated to answer the questions, which are asked regarding images using two sources of information: the entity annotations generated from the scene graphs and the attention map generated by a VQA model. As can be determined from these research works, since scene graphs are able to provide information regarding the relationships between objects in visual scenes, there is significant scope for future research into scene graph-based VQA.

\vspace{-1em}
\subsection{Image Understanding and Reasoning}
Fully understanding an image necessitates the detection and recognition of different visual components, as well as the inference of higher-level events and activities through the combination of visual modules, reasoning modules, and priors.  Therefore, the scene graph with triplets of $\langle subject-relation-object \rangle $ contains information that is essential to image understanding and reasoning. 
Visual understanding requires the model to have visual reasoning ability. However, existing methods tend to pay less attention to how to make a machine (model) "think", and instead attempt to extract the pixel-level features directly from the images; this is despite the fact that it is difficult to carry out accurate reasoning using pixel-level visual features alone. The task of image reasoning should be based directly on the detected objects, rather than on pixel-level visual features. 

More specifically, similar to traditional visual understanding methods, the objects, scenes, and other constituent visual components first need to be detected by a deep learning perception system from input images. A common-sense knowledge base is then built by a Bayesian Network based on the image annotations, while the object interactions are predicted by an intermediate knowledge structure called SDG (Scene Description Graph) \cite{aditya2018image}. These object interaction priors can be used as the input for image reasoning models and applied to other visual reasoning tasks. In addition, we should focus on teaching a machine (model) to ``think'' for visual reasoning tasks, such as, by using XNMS (Explicit and Explicit Neural Modules)\cite{shi2019explainable}. XNMS defines several neural modules that are responsible for specific functions (such as object location, attention transformation, etc.) based on scene graphs. 
XNMS separates "high-level" visual reasoning from "low-level" visual perception and forces the model to focus on how to "think" rather than on simple visual recognition. Since image reasoning is based on object detection and recognition, we hope to learn the mapping from the shared visual feature space by objects and relations to two independent semantic embedding spaces (objects and relations). Moreover, to avoid confusion between these two feature spaces, the visual features of the relationships are not transferred to the objects; instead, only the object features are transferred \cite{zhang2019large}. Visual reasoning based on scene graphs has its applications for reasoning civic issues \cite{kumar2019adversarial}, which are mainly reflected by the relationships between the objects. Furthermore, generating a semantic layout from a scene graph is a crucial intermediate task in the process of connecting textual descriptions to the relevant images.

\vspace{-1em}
\subsection{3D Scene Understanding}
Similar to the 2D scene graph generated from 2D images, a scene graph can also be constructed from 3D scenes as a 3D scene graph, which can provide numerically accurate quantification of the object relationships in 3D scenes. A 3D scene graph succinctly describes the environments by abstracting the objects and their relationships in 3D space in the form of graphs. The construction of a 3D scene is very helpful for the understanding of the indoor complex environment and other tasks.

To construct a 3D scene graph, it is necessary to locate the different objects, identify the elements, attributes, and relationships between the objects in 2D images, and then use all of this information to construct a 3D scene. 
In \cite{armeni20193d} and \cite{kim20203d}, the basic process used to generate 3D scene graphs is similar, and there are several similar methods (Faster RCNN or Mask RCNN) used to extract the required information from a number of 2D images. However, there are differences in the specific details. For example, different methods have been proposed for constructing 3D scene graphs using the relevant information obtained from 2D images. Specifically, in \cite{armeni20193d}, Armeni et al. tried to construct a 3D scene graph of a building. The constructed 3D scene graph consists of four layers: the building, rooms, objects, and cameras. In each layer, there are a set of nodes with their attributes and edges representing the relationships between nodes. Moreover, in \cite{kim20203d}, Kim et al. proposed the 3D scene graph to promote the intelligent agents gathering the semantics within the environments, then applying the 3D scene graph to other downstream tasks. 
Furthermore, the applicability of the 3D scene graph \cite{kim20203d} is verified by demonstrating two major applications of VQA (Visual Question and Answering) and task planning, achieving better performance than the traditional 2D scene graph-based methods. 
Similarly, 3DSSG (3D Semantic Scene Graphs) \cite{wald2020learning} and 3D-DSG (3D Dynamic Scene Graphs) \cite{rosinol2021kimera} also studied the scene understanding of indoor 3D environments. More specifically, 3DSSG proposes a learning method based on PointNet and GCN that moves from the scene point cloud regression to the scene graph. This method has achieved good performance in the 3D scene retrieval task. 
3D-DSG attempts to narrow the perception gap between robots and humans in a 3D environment by capturing the metrics and semantics of the dynamic environment. These works have effectively deepened people's understanding of 3D scenes and promoted related applications.

\vspace{-1em}
\subsection{Human-Object / Human-Human Interaction}
\label{sec:HOI/HHI}
There are many fine-grained categories of things in the scene, which can be generally divided into humans and objects. Therefore, some scene graph-related research works have focused on the detection and recognition of HOI (Human-Object Interaction) \cite{li2018transferable,zhuang2018hcvrd,shen2018scaling,chao2018learning,gkioxari2018detecting} and HHI (Human-Human Interaction) \cite{li2018visual,zhang2019multi} in scenes. In these works, the long tail of relationships remains a problem to be solved \cite{zhuang2018hcvrd,shen2018scaling}, while the detection and recognition of interpersonal relationships have also been proposed \cite{li2018visual,zhang2019multi}; these human-human relationships can be used to further infer the visual social relationships in a scene. In this section, we will discuss the existing method-based scene graph for the detection and recognition of human-object interactions and human-human interaction.

For HOI, there are two main benchmarks: HICO-DET \cite{chao2015HICO} and HCVRD \cite{zhuang2018hcvrd}. This type of visual relational HOI dataset has a natural long-tail distribution and has one-shot or zero-shot HOI detection, which makes it very difficult to conduct model training for most HOI methods in order to achieve better performance. In addition, the task of HOI relies on object detection and involves the construction of human and object pairs with high complexity \cite{li2018transferable}. The zero-shot learning approach is introduced to address the challenges of scaling HOI recognition to the long tail of categories in the HOI dataset \cite{shen2018scaling}. In addition, HOI recognition is an important means of distinguishing between different types of human actions that occur in the real world. Most HOI methods consist of two steps: human-object pair detection and HOI recognition \cite{xu2019learning,li2018transferable}. The detected proposals of paired human-object regions are passed into a multi-stream network (HO-RCNN \cite{xu2019learning} and iCAN \cite{li2018transferable}) to facilitate the classification of HOIs by extracting the features from the detected humans, objects, and the spatial relations between them. Moreover, the structural knowledge from the images is also beneficial for HOI recognition, with GCN being a commonly used model for learning structural features. For example, GPNN (Graph Parsing Neural Network) is proposed in \cite{qi2018learning} to infer the HOI graph structure represented by an adjacency matrix and node labels. Furthermore, in order to reduce the number of human-object pairs, some inter-activeness priors can be explored for HOI detection; these indicate whether a human and object have interactions with each other \cite{li2018transferable}, and can be learned from the HOI datasets, regardless of HOI category settings.

The above HOI approaches focus primarily on the detection, selection, and recognition of human-object pairs. However, they do not consider whether the approach adopted for the corresponding HOI tasks should be human-centric or object-centric. In a given scene, however, most human-object interactions are human-centric. Therefore, some HOI works have adopted human-centric approaches such as human-to-object \cite{chao2018learning,gkioxari2018detecting} and human-to-human \cite{li2018visual,zhang2019multi}. Inspired by a human-centric approach, we can first identify a human in a scene, then select the human-object pairs of interest to facilitate the recognition of human-object pairs in multi-stream networks; of these, HO-RCNN \cite{chao2018learning} is a representative example. In addition, the information of HOI can be used for action recognition. InteractNet\cite{gkioxari2018detecting} may be the first proposed multi-task network for human-centric HOI detection and action recognition. This network model can achieve the task of detecting $\langle human-verb-object \rangle $ triplets in challenging images. 
Moreover, it was hypothesized that the visual features of the detected persons contain powerful cues for localizing the objects with which they interact so that the model learns to predict the action-specific density over the object locations based on the visual features of the detected persons.

Furthermore, interactions can also occur between humans in a scene, which indicate social relationships. The identification of social relationships in a scene requires a deeper understanding of the scene, along with a focus on human-to-human rather than human-to-object interaction. Therefore, social relationship detection is a task of human-centric HHI, and related works mainly consist of human-human pair detection and social relationship recognition using two network branches \cite{li2018visual,zhang2019multi}. For social relationship recognition, contextual cues can be exploited by a CNN model with attention mechanisms \cite{li2018visual}. Adaptive Focal Loss is designed by leveraging the ambiguous annotations so that the model can more effectively learn the relationship features; the goal here is to solve the problem of uncertainty arising during the visual identification of social relationships. The global visual features and mid-level details are also beneficial for social relationship recognition, and GCN is a commonly used model for predicting human social relationships by integrating the global CNN features \cite{zhang2019multi}.

\vspace{-1em}
\section{Datasets and performance evaluation}
\label{sec:Datasets and performance evaluation}
\subsection{Datasets}
\label{sec:SGG_Datasets}

In this section, we present a detailed summary of the datasets commonly used in SGG tasks, so that interested readers can make their selection accordingly. 
We investigated a total of 14 commonly used datasets, including 10 static image scene graph datasets, 2 video scene graph datasets and 2 3D scene graph datasets.
\begin{itemize}
    \item \textit{RW-SGD} \cite{johnson2015image} is constructed by manually selecting 5,000 images from YFCC100m \cite{thomee2015new} and Microsoft COCO datasets \cite{lin2014microsoft}, after which AMT (Amazon’s Mechanical Turk) is used to produce a human-generated scene graph from these selected images.
    
    \item \textit{Visual Relationship Dataset} (VRD) \cite{lu2016visual} is constructed for the task of visual relationship prediction. The construction of VRD highlights the long tail of infrequent relationships.

    \item \textit{UnRel Dataset} (UnRel-D) \cite{peyre2017weakly} is a new challenging dataset of unusual relations, and contains more than 1,000 images, which can be queried with 76 triplet queries. The relatively small amount of data and visual relationships in UnRel-D make the long-tail distribution of relationships in this dataset not obvious.

    \item \textit{HCVRD Dataset} \cite{zhuang2018hcvrd} contains 52,855 images with 1,824 object categories and 927 predicates, along with 28,323 relationship types. Similar to VRD, HCVRD also has a long-tail distribution of infrequent relationships. 

    \item \textit{Open Images} \cite{xu2017scene} is a large-scale dataset, which provides a large number of examples for object detection and relationship detection.
    
    \item \textit{Visual Phrase} \cite{Sadeghi2011Recognition} is a dataset that contains visual relationships, and it is mainly used to improve object detection. It contains 13 common types of relationships.
    
    \item \textit{Scene Graph} \cite{schroeder2020structured} is a dataset containing visual relationships, which is designed to improve image retrieval tasks. Although it contains 23,190 relationship types, it has only 2.3 predicates per object category.
    
    \item \textit{Visual Genome} (VG) \cite{krishna2017visual}, \textit{VG150} \cite{xu2017scene} and \textit{VrR-VG} (Visually-Relevant Relationships Dataset) \cite{liang2019vrr}. VG is a large-scale visual dataset consisting of various components, including objects, attributes, relationships, question-answer pairs, and so on.
    VG150 and VrR-VG are two datasets constructed based on VGD. VG150 uses VGD to eliminate objects with poor annotation quality, overlapping bounding boxes, and/or ambiguous object names, and retains 150 commonly used object categories.
    VrR-VG is constructed based on VGD by filtering out visually irrelevant relationships. The top 1,600 objects and 500 relationships are selected from VG by applying a hierarchical clustering algorithm on the relationships’ word vectors. Therefore, VrR-VG is a scene graph dataset used to highlight visually relevant relationships; the long-tail distribution of relationships is largely suppressed on this dataset.
    
    \item \textit{CAD120++} \cite{zhuo2019explainable} and \textit{Action Genome} \cite{ji2019action} are two video action reasoning datasets containing scenes of human daily life. They can be used for the analysis of tasks related to spatio-temporal scene graphs.
    
    \item \textit{Armeni et al.} \cite{armeni20193d} and \textit{3DSSG} \cite{wald2020learning} are two large-scale 3D semantic scene graphs containing indoor architecture or 3D reconstructions of real scenes. They are widely used as part of research in areas related to 3D scene understanding (robot navigation, augmented and virtual reality, etc.).
\end{itemize}
The information of these datasets is summarized in TABLE \ref{tab:datasets}. This includes various attributes of datasets commonly used in SGG tasks. 

\vspace{-1em}
\subsection{Evaluation method and performance comparison}
\label{sec:Evaluation method and performance comparison}
\begin{figure} 
\center{\includegraphics[width=0.49\textwidth] {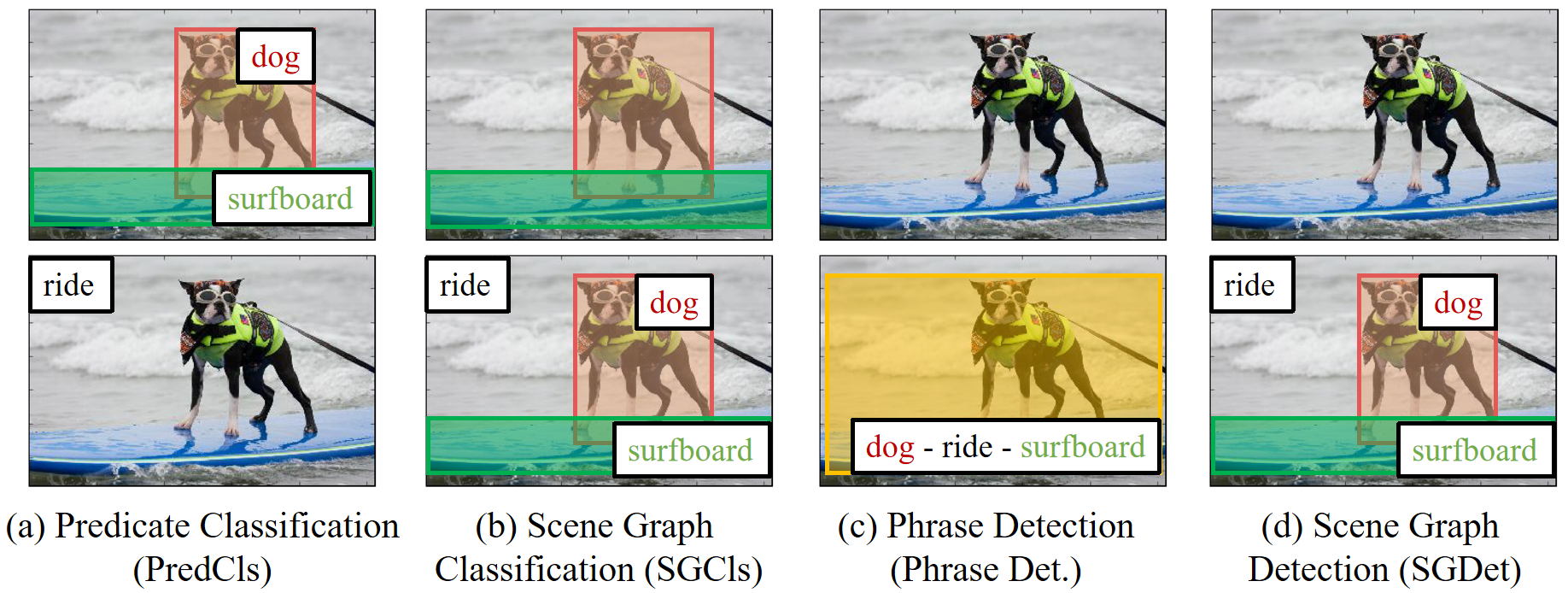}}
\vspace{-2 em}
\caption{Four commonly used evaluation methods in visual relationship detection.}
\vspace{-1.5 em}
 \label{fig:Evaluation1}
\end{figure}

Visual relationship detection is the core of SGG. The commonly used evaluation methods for visual relationship detection are as follows \cite{lu2016visual}:
\begin{itemize}
    \item \textit{Predicate detection (Predicate Det.)} (Fig.\ref{fig:Evaluation1}(a)). It is required to predict possible predicates between object pairs given a set of localized objects and object labels. The goal is to study relationship prediction performance without being restricted by object detection \cite{girshick2014rich}.
    \item \textit{Phrase detection (Phrase Det.)} (Fig.\ref{fig:Evaluation1}(c)). It is required to predict a $\langle subject-predicate-object \rangle$ triplets for a given image and locate a bounding box for the entire relationship that overlaps the ground truth box by at least 0.5 \cite{Sadeghi2011Recognition}.
    \item \textit{Relationship detection (Relationship Det.)} (Fig.\ref{fig:Evaluation1}(d)). It is required to output a set of triplets of a given image, while localizing the objects in the image.
\end{itemize}

On the other hand, the above three visual relationship detection evaluation methods do not consider the long-tail distribution phenomenon and the graph-level coherence that are common in real scenes. For this reason, \cite{Tang2020} further improves the evaluation method of SGG; that is, SGG diagnosis. SGG diagnosis is based on the following three key metrics:
\begin{itemize}
    \item \textit{Relationship Retrieval (RR).} This can be further divided into three sub-tasks. 
    (1) Predicate Classification (PredCls): the same as Predicate Det.. 
    (2) Scene Graph Classification (SGCls) (Fig.\ref{fig:Evaluation1}(b)): its input is a ground true bounding box without labels.
    (3) Scene Graph Detection (SGDet): detects scene graphs from scratch. It is the same as Relationship Det..
    \item \textit{Zero-Shot Relationship Retrieval (ZSRR)} \cite{lu2016visual}. The visual relationship that ZSRR requires to be tested has never been observed in the training set. Similar to RR, ZSRR has the same three sub-tasks.
    \item \textit{Sentence-to-Graph Retrieval (S2GR).} Both RR and ZSRR are evaluated at the triplet level. S2RR aims to evaluate the scene graph at the graph level. It uses image caption sentences as queries to retrieve images represented by scene graphs.
\end{itemize}

\begin{table*}[t]
\centering
\caption{\textcolor{black}{Performance comparison of the SGG method with graph constraints on VG150.} \cite{xu2017scene}. '-' indicates that there is no corresponding result released in the paper. Note that some methods use multiple different types of networks or priors, and these are categorized according to the most prominent method they use. $\dagger$ denotes the re-implemented version from \cite{zellers2018neural}. * indicates that the method evaluates on other datasets. The best performance is shown in bold.}
\vspace{-1 em}
\label{tab:PerformanceComparison}
\resizebox{\textwidth}{!}{
\begin{tabular}{|l|l|l|cc|cc|cc|}
\hline
&&& \multicolumn{2}{c|}{PredCls} & \multicolumn{2}{c|}{SGCls} & \multicolumn{2}{c|}{SGDet} \\
&&Method & R@\{20/50/100\} & mR@\{20/50/100\} & R@\{20/50/100\} & mR@\{20/50/100\} & R@\{20/50/100\} & mR@\{20/50/100\} \\ \hline

\multirow{28}{*}{SGG}&\multirow{4}{*}{TransE}& VTransE* \cite{zhang2017visual} &  -/-/- &  -/-/- &  -/-/- &  -/-/- &  -/5.5/6.0 &  -/-/- \\
&& GPS-Net \cite{lin2020gps} & \textbf{67.6/69.7/69.7} & -/-/22.8 & \textbf{41.8/42.3/42.3} & -/-/12.6 & 22.3/\textbf{28.9/33.2} & -/-/9.8 \\
&&VTransE-TDE \cite{Tang2020} & -/-/- & \textbf{18.9}/25.3/28.4 & -/-/- & \textbf{9.8/13.1}/14.7 & -/-/- & \textbf{6.0/8.5/10.2} \\
&&Motifs-TDE \cite{Tang2020, zellers2018neural} & 33.6/46.2/51.4 & 18.5/\textbf{25.5/29.1}  & 21.7/27.7/29.9 & \textbf{9.8/13.1/14.9} & 12.4/16.9/20.3& 5.8/8.2/9.8 \\
&&MemNet \cite{wang2019exploring} & 42.1/53.2/57.9 & -/-/- & 23.3/27.8/29.5 & -/-/- & 7.7/11.4/13.9 & -/-/- \\
&&LSBR \cite{he2020learning}& 60.3/66.2/68.0 & -/-/- & 33.6/37.5/38.3 & -/-/- &  \textbf{23.6}/28.2/31.4  & -/-/- \\ 
\cline{2-9}

&\multirow{5}{*}{CNN}& Pixels2Graphs \cite{newell2017pixels} & 47.9/54.1/55.4 & -/-/- & 18.2/21.8/22.6 & -/-/- & 6.5/8.1/8.2 & -/-/- \\
&&Graph R-CNN \cite{yang2018graph} & -/54.2/59.1 & -/-/- & -/29.6/31.6 & -/-/- & -/11.4/13.7 & -/-/- \\
&&HCNet \cite{ren2020scene} & 59.6/66.4/68.8 & -/-/- & 34.2/36.6/\textbf{37.3} & -/-/- & \textbf{22.6}/28.0/31.2 & -/-/- \\
&&Assisting-SGG \cite{Inuganti2020AssistingSG}& \textbf{67.2/68.9/68.9} & -/-/- & \textbf{36.4/37.0}/37.0 & -/-/- & 21.7/\textbf{28.3/32.6}  & -/-/- \\ 
&&\textcolor{black}{FCSGG}  \cite{liu2021fully}& 33.4/41.0/45.0 & \textbf{4.9/6.3/7.1} & 19.0/23.5/25.7 & \textbf{2.9/3.7/4.1} & 16.1/21.3/25.1  & \textbf{2.7/3.6/4.2} \\
\cline{2-9}

&\multirow{11}{*}{\makecell{RNN/\\LSTM}} & IMP \cite{xu2017scene} & -/44.8/53.0 & -/6.1/8.0 & -/21.7/24.4 & -/3.1/3.8 & -/3.4/4.2 & -/0.6/0.9 \\ 
&&\textcolor{black}{IMP$\dagger$} \cite{xu2017scene, zellers2018neural} & 58.5/65.2/67.1 & -/9.8/10.5 & 31.7/34.6/35.4 & -/5.8/6.0 & 14.6/20.7/24.5  & -/3.8/4.8 \\
&&FREQ \cite{zellers2018neural, tang2019learning} & 49.4/59.9/64.1 & 8.3/13.0/16.0 & 27.7/32.4/34.0 & 5.1/7.2/8.5 & 17.7/23.5/27.6  & 4.5/6.1/7.1 \\
&&FREQ+Overlap \cite{zellers2018neural} &  53.6/60.6/62.2  & -/-/- & 29.3/32.3/32.9 & -/-/- & 20.1/26.2/30.1   &-/-/- \\
&&VCTree-SL \cite{tang2019learning} & 59.8/66.2/67.9 & -/-/- & 35.0/37.9/38.6 & -/-/- & 21.7/27.7/31.1 & -/-/- \\
&&VCTree-HL \cite{tang2019learning} & 60.1/66.4/68.1 & 14.0/17.9/19.4 & 35.2/38.1/38.8 & 8.2/10.1/10.8 & 22.0/27.9/31.3 & 5.2/6.9/8.0 \\
&&VCTree L2+cKD \cite{Wang2020TacklingTU}& 59.0/65.4/67.1 & 14.4/18.4/20.0 & 41.4/45.2/46.1 & 9.7/12.1/13.1 & 24.8/32.0/36.1  & 5.7/7.7/9.1 \\ 
&&VCTree-TDE$_{Cross Entropy}$ \cite{suhail2021energy} & -/-/- &16.3/22.85/26.26& -/-/- &11.85/15.81/17.99& -/-/- &6.59/8.99/10.78 \\
&&VCTree-TDE$_{EBM\_Loss}$ \cite{suhail2021energy}& -/-/- &\textbf{19.87/26.66/29.97}& -/-/- &\textbf{13.86/18.2/20.45}& -/-/- &\textbf{7.1/9.69/11.6} \\
&&SIG \cite{wang2020sketching}& 59.8/66.3/68.1 & -/-/- & 33.8/36.6/37.3 & -/-/- & 21.6/27.5/30.9 & -/-/- \\
&&PANet \cite{chen2019panet}&  59.7/66.0/67.9 & -/-/- & 37.4/40.9/41.8 & -/-/- & 21.5/26.9/29.9  & -/-/- \\
&&BGT-Net \cite{dhingra2021bgt}&  \textbf{60.9/67.1/68.9} & -/-/- & \textbf{41.7/45.9/47.1} & -/-/- & \textbf{25.5/32.8/37.3}  & -/-/- \\ 
\cline{2-9}

&\multirow{13}{*}{Graph}&CMAT \cite{chen2018scene} & 60.2/66.4/68.1 & -/-/- & 35.9/\textbf{39.0/39.8} & -/-/- & \textbf{22.1}/27.9/31.2 & -/-/- \\
&&CMAT-XE \cite{chen2019counterfactual} & -/-/- & -/-/- & 34.0/36.9/37.6 & -/-/- & -/-/- & -/-/- \\
&&RelDN \cite{zhang2019graphical} & \textbf{66.9}/68.4/68.4 & -/-/- & \textbf{36.1}/36.8/36.8 & -/-/- & 21.1/28.3/32.7 & -/-/- \\
&&MSDN \cite{li2017scene} & -/67.0/\textbf{71.0} & -/-/- & -/-/- & -/-/- & -/10.7/14.2 & -/-/- \\
&&FactorizableNet* \cite{li2018factorizable} & -/22.8/28.6 & -/-/- & -/-/- & -/-/- & -/13.1/16.5 & -/-/- \\ 
&&ARN \cite{qi2019attentive}& -/65.1/66.9 & -/-/- & -/36.5/38.8 & -/-/- & -/-/-  & -/-/- \\ 
&&VRF \cite{dornadula2019visual}& -/56.7/57.2 & -/-/- & -/23.7/24.7 & -/-/- & -/13.2/13.5  & -/-/- \\ 
&&HOSE-Net \cite{Wei2020HOSENetHO}& -/\textbf{70.1}/70.1 & -/-/- & -/ 37.3/37.3& -/-/- & -/\textbf{28.9/33.3} & -/-/- \\ 
&&IMP+GLAT+Fusion \cite{zareian2020learning}& -/-/- & -/12.1/12.9 & -/-/- & -/6.6/ 7.0 & -/-/-  & -/-/- \\ 
&&SNM+GLAT+Fusion \cite{zareian2020learning}& -/-/- & -/14.1/15.3 & -/-/- & -/7.5/7.9 & -/-/-  & -/-/- \\ 
&&KERN+GLAT+Fusion \cite{zareian2020learning}& -/-/- & -/17.8/19.3 & -/-/- & -/9.9/10.4 & -/-/-  & -/-/- \\ 
&&Dual-ResGCN \cite{Zhang2020DualRF}& 60.2/66.6/68.2 & \textbf{15.6/19.7/21.5} & 35.4/38.3/39.1 & \textbf{9.1/11.1/12.0} & \textbf{22.1}/28.1/31.5  & \textbf{6.1/8.4/9.5} \\
&&\textcolor{black}{RAAL-SGG} \cite{liu2021toward}& 59.1/66.2/68.4 & 14.4/18.3/19.9  & 33.5/36.7/37.6 & 7.9/9.6/10.3 & 21.7/27.3/29.9  & 4.9/6.5/7.4\\
\hline

\multirow{12}{*}{\makecell[l]{Prior\\knowledge}}&\multirow{2}{*}{\makecell[l]{Language\\Prior}} & SG-LP \cite{lu2016visual} & -/27.9/35.0 & -/-/- & -/11.8/14.1 & -/-/- & -/0.3/0.5 & -/-/- \\
&&TFR \cite{jae2018tensorize}& \textbf{40.1/51.9/58.3} & -/-/- & \textbf{19.6/24.3/26.6} & -/-/- & \textbf{3.4/4.8/6.0}  & -/-/- \\
\cline{2-9}

&\multirow{4}{*}{\makecell[l]{Statistical\\Prior}}&KERN \cite{chen2019knowledge}& \textbf{59.1/65.8}/67.6 & -/17.7/19.2 & 32.3/\textbf{36.7}/37.4 & -/9.4/10.0 & \textbf{22.3}/27.1/29.8  & -/6.4/7.3 \\
&&MotifNet \cite{zellers2018neural, tang2019learning} & 58.5/65.2/67.1 & \textbf{10.8}/14.0/15.3 & \textbf{32.9}/35.8/36.5 & \textbf{6.3}/7.7/8.2 & 21.4/\textbf{27.2}/30.3 & \textbf{4.2}/5.7/6.6 \\
&&MotifNet-freq  \cite{zellers2018neural} & -/41.8/48.8 & -/-/- & -/23.8/27.2 & -/-/- & -/6.9/9.1 & -/-/- \\
&&\textcolor{black}{ResCAGCN + PUM} \cite{yang2021probabilistic} & -/-/\textbf{68.3} & -/\textbf{20.2/22.0} & -/-/\textbf{39.0} & -/\textbf{11.9/12.8} & -/-/\textbf{31.3} & -/\textbf{7.7/8.9} \\
\cline{2-9}

&\multirow{7}{*}{\makecell[l]{Knowledge\\Graph}} & IMP+ \cite{chen2019knowledge, xu2017scene} & \textbf{52.7}/59.3/61.3 & -/9.8/10.5 & \textbf{31.7}/34.6/35.4 & -/5.8/6.0 & \textbf{14.6}/20.7/24.5 & -/3.8/4.8 \\
&&GPI \cite{herzig2018mapping} & -/65.1/66.9 & -/-/- & -/36.5/38.8 & -/-/- & -/-/- & -/-/- \\
&&KB-GAN \cite{gu2019scene} & -/-/- & -/-/- & -/-/- & -/-/- & -/13.7/17.6 & -/-/- \\ 
&&GB-Net \cite{zareian2020bridging}& -/66.6/68.2 & -/\textbf{19.3/20.9} & -/38.0/38.8 & -/9.6/10.2 & -/26.4/30.0  & -/\textbf{6.1}/\textbf{7.3} \\ 
&&DG-PGNN \cite{khademi2020deep}& -/\textbf{70.1/73.0} & -/-/- & -/\textbf{39.5/40.8} & -/-/- & -/\textbf{32.1/33.1}  & -/-/- \\ 
&&Schemata \cite{Sharifzadeh2021ClassificationBA}& -/66.9/68.4 & -/19.1/20.7 & -/39.1/39.8 & -/\textbf{10.1/10.9} & -/-/-  & -/-/- \\ 
\hline
\end{tabular}
}
\vspace{-1 em}
\end{table*}

Recall@K (R@K) is often used as an evaluation metric for the above tasks. In addition, due to the existence of reporting bias in R@K \cite{misra2016seeing}, R@K is easily disturbed by high-frequency predicates. Therefore, mean Recall@K (mR@K) \cite{chen2019knowledge, tang2019learning} was proposed. mR@K retrieves each predicate separately and then averages R@K for all predicates. 
Graph constraints are also a factor that should be considered. Some previous work \cite{xu2017scene} constrained only one relationship for a given object pair when calculating R@K, while other work \cite{newell2017pixels, chen2019knowledge} omitted this constraint and enabled multiple relationships to be obtained. In addition, to better evaluate the model's ability to conduct rare visual relationship prediction, \cite{knyazev2020graph} introduces density-normalized edge loss and explores a novel weighted metric.

At present, most existing SGG methods use three subtasks in \textit{RR} with graph constraints for performance evaluation. Referring to the classification outlined in Section \ref{sec:SGG_Alone} and Section \ref{sec:SGG_Proir}, the performance of the related SGG methods are summarized in TABLE \ref{tab:PerformanceComparison}. Most current methods use the SGG method based on graph and RNN/LSTM. Compared with R@K, mR@K is generally lower. For datasets with obvious long-tail distribution, mR@K is a fairer evaluation metric. 
\textcolor{black}{In particular, due to VTransE* \cite{zhang2017visual}, FactorizableNet* \cite{li2018factorizable}, IMP \cite{xu2017scene}, IMP$\dagger$ \cite{xu2017scene, zellers2018neural}, Pixels2Graphs \cite{newell2017pixels}, FCSGG \cite{liu2021fully}, and VRF \cite{dornadula2019visual}, only visual features are used, and the performance of these methods is generally low. In contrast, VCTree \cite{tang2019learning}, KERN \cite{chen2019knowledge}, GPS-Net \cite{lin2020gps}, GB-NET \cite{zareian2020bridging}, in addition to using visual features, they also use other knowledge (for example, language embedding, statistical information, and counterfactual causality, etc.). This allows these methods to gain more additional knowledge and thus more performance gains. In addition, currently almost all methods propose that the prediction of objects and relationships is associated instead of independent. They try to consider the contextual information between objects \cite{newell2017pixels, yu2017visual, yin2018zoom, zhang2018interpretable}, or use GNN \cite{li2017scene,li2018factorizable,qi2019attentive}, LSTM \cite{chen2019panet,zellers2018neural,tang2019learning}, message passing \cite{xu2017scene} and other methods to capture this relationship. However, these methods often use cross entropy loss training, as shown in Eq.\ref{eq:EBM}, which is essentially considering objects and relationships as independent entities. A new energy-based loss function is proposed in EBM \cite{suhail2021energy}. This well-designed loss calculates the joint likelihood of objects and relationships. Compared with the traditional cross-entropy loss, it has achieved a consistent improvement in mR@K on multiple classic models \cite{Tang2020, tang2019learning, xu2017scene}. At the same time, EBM \cite{suhail2021energy} has achieved the best performance among the algorithms currently investigated.}
In addition, ZSRR is also an important task, but most current methods fail to evaluate the ZSRR task. Paying attention to the evaluation of the ZSRR task is helpful to the study of the long-tail distribution of the scene graph.

\vspace{-1em}
\section{Future Research}
\label{sec:FutureDirection}

SGG aims at mining the relationships between objects in images or scenes and forming relationship graphs. Although SGG currently has a lot of related research, there are still many directions worthy of attention.

\textbf{Long-tailed distribution in visual relations.} The inexhaustibility of visual relations and the complexity of real scenes determine the inevitability of long-tailed distribution in visual relations. The data balance required by model training and the long-tailed distribution in real data presents a pair of unavoidable contradictions. Associative reasoning through similar objects or similar relationships across scenes may be a suitable research direction to pursue, as it may aid in solving the long-tail distribution problem of relationships on the current scene graph dataset to a certain extent. In addition, a targeted long-tail distribution evaluation metric or task design is also a potential research direction, as it could aid in fairer evaluation of the model's learning ability in zero/one/few-shot contexts; however, related research is still very limited. Although the long-tail problem has received extensive attention from researchers (Section \ref{sec:long-tail}), there are large numbers of potential, unfrequent, non-focused, or even unseen relationships in the scene that still need to be explored.

\textbf{Relationships detection between distant objects.} Currently, a scene graph is generated based on large numbers of small-scale relationship graphs, which are abstracted from small scenes in scene graph datasets by means of relevant relationship prediction and reasoning models. The selection of potential effective relationships \cite{kolesnikov2019detecting,zhang2017relationship} and the establishment of the final relationships in the scene graph are largely dependent on the spatial distance between objects, such that no relationship will exist between two distant objects. However, in the case of a large scene, there are still more such relationships \cite{zhang2019large}. Therefore, an appropriate proportion of large-scale images can be added to the existing scene graph datasets, while the relationships between objects separated by long distances can be properly considered during SGG, which will improve the integrity of the scene graph.

\textbf{SGG based on dynamic images.} The scene graph is generated based on static images in scene graph datasets, and object relationship prediction is also carried out for the static objects in the images by means of related reasoning models. 
There are very few related research works \cite{aditya2018image,shi2019explainable,ji2019action}, and little attention has been paid to the role played by the dynamic behaviors of objects in the prediction and inference of the relationships.
In practice, however, it may be necessary to predict large numbers of relationships by means of successive actions or events; that is, relationship detection and reasoning based on video scenes. This is because, compared with static images, it is obvious that the analysis of spatio-temporal scene graphs of dynamic images offers a wider range of application scenarios. 
We therefore contend that it will be necessary to focus on relationship prediction based on the dynamic actions of the objects in the video.

\textbf{Social relationship detection based on scene graph.} From Section \ref{sec:HOI/HHI}, we can see that the detection of HOI (human-object interaction) and human-human interaction is an important application of scene graphs, and these types of relationships can be further extended to detect social relationships. We believe that social relationship detection can be used to understand the scenes in more depth, and is thus also a very important research direction. The SGG models based on large-scale datasets can even mine unseen social relationships from the visual data, which yields a wider range of practical application values.

\textbf{Models and methods of visual reasoning.} For SGG, the mainstream methods have been developed based on object detection and recognition.
However, due to the limitations in the current scene graph datasets and the limited capability of relationship prediction models derived using these datasets, it is difficult for existing models to continuously enhance their ability to predict relationships. Therefore, we believe that online learning, reinforcement learning, and active learning may be relevant methods or strategies that could be introduced into future SGG methods, as this would enable the SGG models to continuously enhance their relationship prediction abilities by drawing on a large number of constantly updated realistic datasets.

In general, research in the field of scene graphs has developed rapidly and has broad application prospects. Scene graphs are expected to further promote the understanding and reasoning of higher-level visual scenes. At present, however, scene graph-related research is not sufficiently mature, meaning that it requires more effort and exploration.

\vspace{-1em}
\section{Conclusion}
\label{sec:Conclusion}
As a powerful tool for high-level understanding and reasoning analysis of scenes, scene graphs have attracted an increasing amount of attention from researchers. However, research into scene graphs is often cross-modal, complex, and rapidly developing. At the same time, no comprehensive review of scene graph-related research could be found at the time of writing. For this reason, we conducted a comprehensive and systematic survey of SGG. In particular, we classified existing SGGs based on the SGG method and the introduction of additional prior knowledge. We then conducted a comprehensive survey of the application of SGG. In addition, we presented detailed statistics on the datasets used in the context of the scene graph to aid interested readers in selecting their preferred approach. Finally, we discussed in detail the future development directions of the scene graph. Therefore, we have reason to believe that this survey will be very helpful for expanding readers' understanding of scene graph development and related research.
\vspace{-1em}

\ifCLASSOPTIONcompsoc
  \section*{Acknowledgments}
\else
  \section*{Acknowledgment}
\fi

This work was partially supported by Australian Research Council (ARC) Discovery Early Career Researcher Award (DECRA) under grant no. DE190100626, partially supported by National Natural Science Foundation of China under grant agreements Nos. 61973250, 62073218 and 61906109, partially supported by Natural Science Outstanding Youth Fund of Shandong Province [ZR2021YQ44], partially supported by Shaanxi Provincial Department of Education serves local scientific research under 19JC038, and partially supported by The Key Research and Development Program of Shaanxi under 2021GY-077.


%

\ifCLASSOPTIONcaptionsoff
  \newpage
\fi




\bibliographystyle{IEEEtran}
\bibliography{Bib}

%


\begin{IEEEbiography}[{\includegraphics[width=0.9in,height=1.26in]{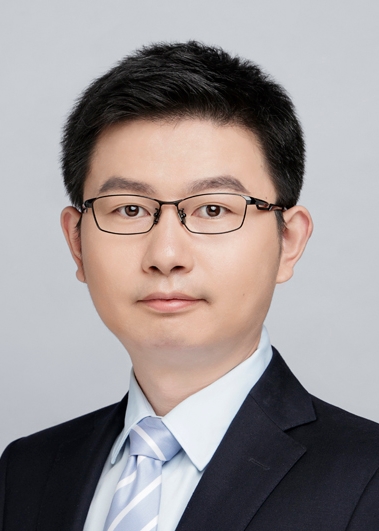}}]{Xiaojun Chang} is a Professor at Faculty of Engineering and Information Technology, University of Technology Sydney.
He was an ARC Discovery Early Career Researcher Award (DECRA) Fellow between 2019-2021. After graduation, he was worked as a Postdoc Research Associate in School of Computer Science, Carnegie Mellon University, a Senior Lecturer in Faculty of Information Technology, Monash University, and an Associate Professor in School of Computing Technologies, RMIT University. He mainly worked on exploring multiple signals for automatic content analysis in unconstrained or surveillance videos and has achieved top performance in various international competitions. He received his Ph.D. degree from University of Technology Sydney. His research focus in this period was mainly on developing machine learning algorithms and applying them to multimedia analysis and computer vision.
\end{IEEEbiography}

\begin{IEEEbiography}[{\includegraphics[width=0.9in,height=1.26in]{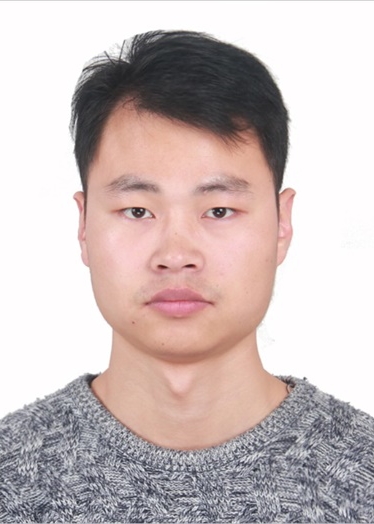}}]{Pengzhen Ren}
is currently studying for a PhD degree at School of Information Science and Technology, Northwest University, Xi'an, China. His main research interests include: neural architecture search, cross-modal retrieval, and multimedia data mining. Currently, he is mainly focused on video action recognition with neural architecture search.
\end{IEEEbiography}

\begin{IEEEbiography}[{\includegraphics[width=0.9in,height=1.26in]{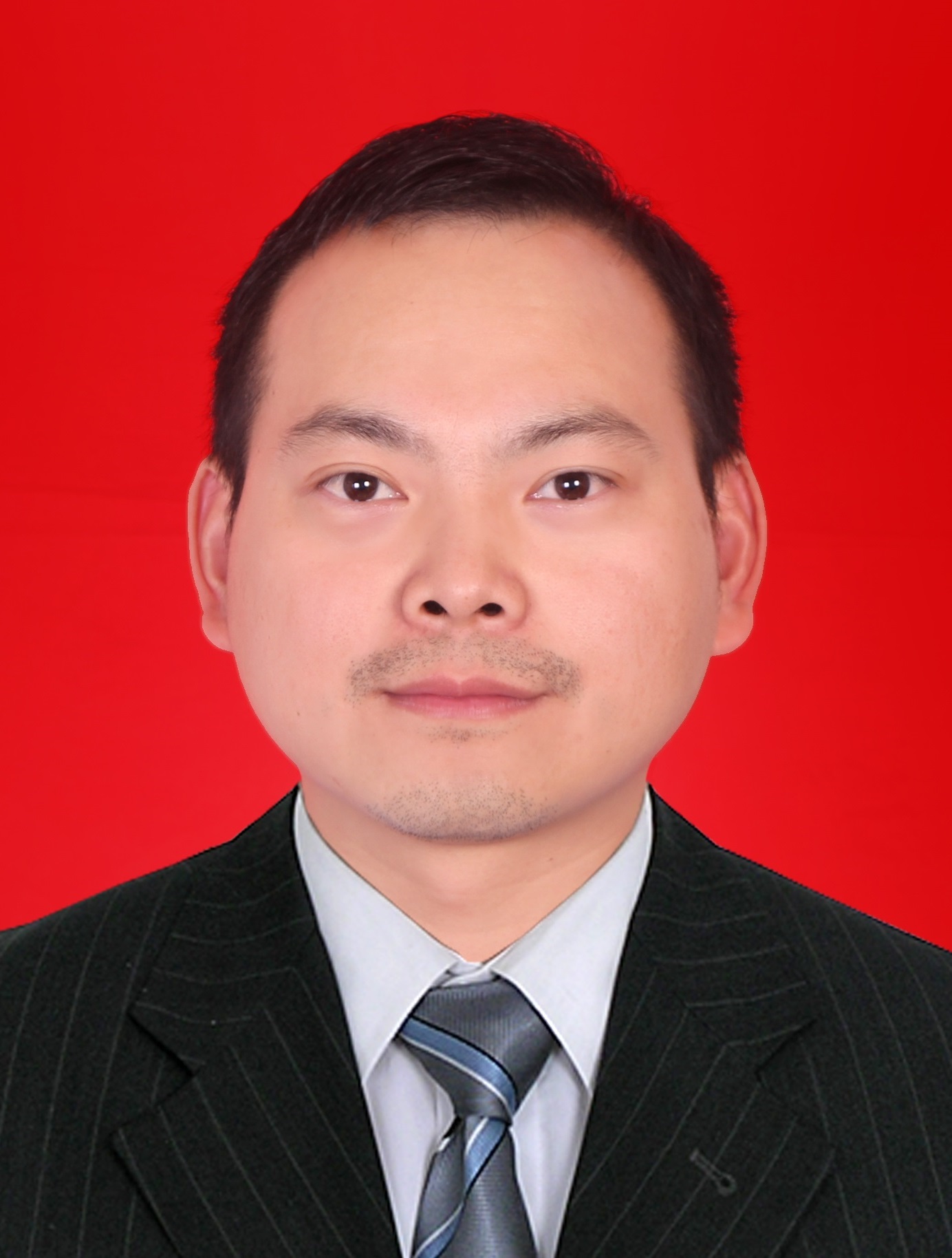}}]{Pengfei Xu} 
is an associate professor at School of Information Science and Technology at Northwest University, Xi’an, China, received the Ph.D. degree from Xidian University in 2014. He has spent most of his time working on computer vision and pattern recognition, and has published more than 50 papers in international journals and conferences, including IEEE TIP, IEEE TGRS and IJCAI etc. He won the first prize of Shaanxi Provincial Natural Science Award in 2020. His research on facial recognition and behavior analysis of Golden Monkey was interviewed by CCTV-13.
\end{IEEEbiography}

\begin{IEEEbiography}[{\includegraphics[width=1in,height=1.25in,clip,keepaspectratio]{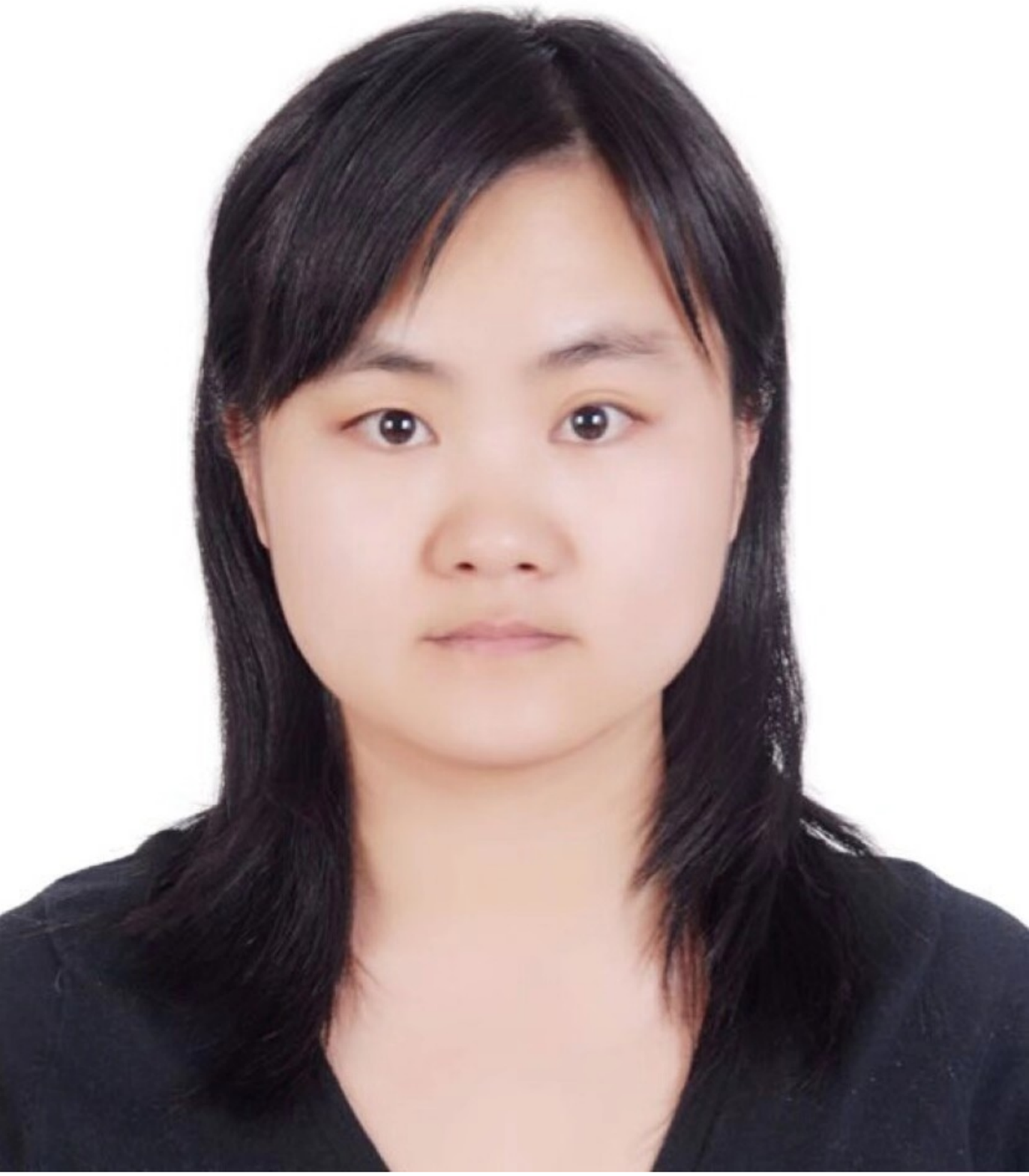}}]{Zhihui Li} is a Research Professor with Shandong Artificial Institute, Qilu University of Technology. Before joined SDAI, she obtained her PhD degree from School of Computer Science and Engineering, University of New South Wales (UNSW). Her research interests are machine learning and its applications to computer vision and multimedia. She has published more than 40 high-quality papers on top-tier venues, three of which are identified as ESI highly cited papers. \end{IEEEbiography}

\begin{IEEEbiography}[{\includegraphics[width=0.9in,height=1.26in]{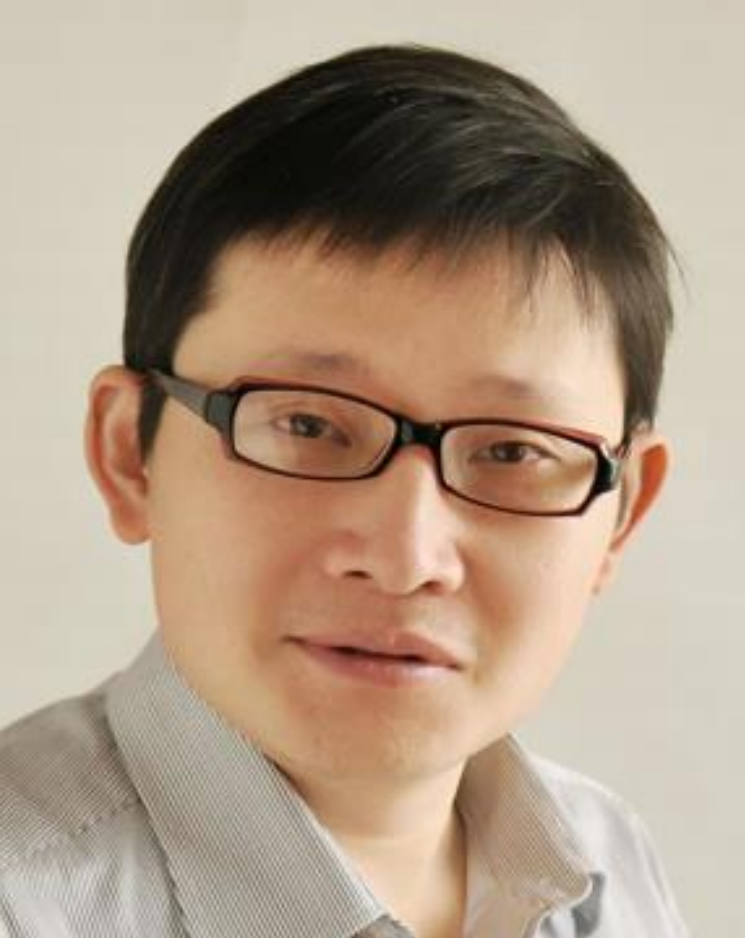}}]{Xiaojiang Chen}
received the Ph.D. degree in computer software and theory from Northwest University, Xi’an, China, in 2010. He is currently a Professor with the School of Information Science and Technology, Northwest University. His current research interests include localization and performance issues in wireless ad hoc, mesh, sensor networks and machine learning. 
\end{IEEEbiography}

\begin{IEEEbiography}[{\includegraphics[width=1in,height=1.25in,clip,keepaspectratio]{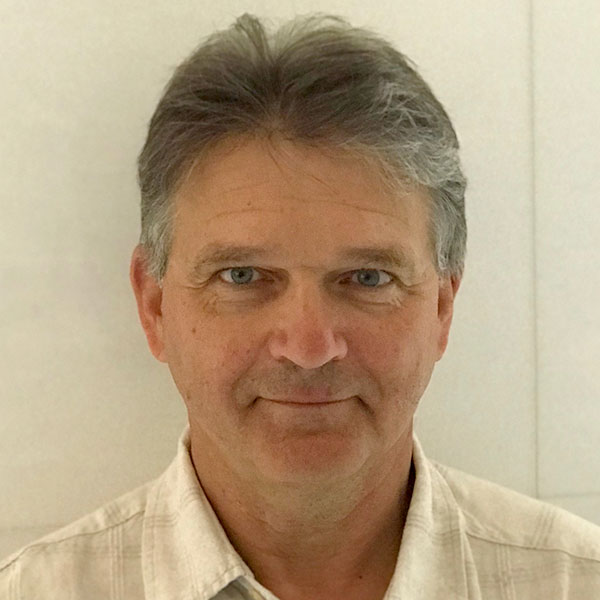}}]{Alex Hauptmann}
received the B.A. and M.A. degrees in psychology from Johns Hopkins University, Baltimore, MD, USA, the Ph.D. degree in computer science from the Technische Universitat Berlin, Berlin, Germany, in 1984, and the Ph.D. degree in computer science from Carnegie Mellon University (CMU), Pittsburgh, PA, USA, in 1991. From 1984 to 1994, he worked on speech and machine translation, when he joined the Informedia Project for digital video analysis and retrieval, and led the development and evaluation of news-on-demand applications. He is currently with the Faculty of the Department of Computer Science and the Language Technologies Institute, CMU. His current research interests include man-machine communication, natural language processing, speech understanding and synthesis, video analysis, and machine learning.
\end{IEEEbiography}




\end{document}